\documentclass[journal]{IEEEtran}
\usepackage{amsmath,amsfonts}
\usepackage{array}
\usepackage{textcomp}
\usepackage{stfloats}
\usepackage{url}
\usepackage{verbatim}
\usepackage{graphicx}
\usepackage{cite}
\usepackage{xcolor}

\usepackage{booktabs}

\usepackage{mathtools}

\usepackage{amssymb}

\usepackage{bm}

\usepackage{tikz}
\usepackage{pgfplots}
\pgfplotsset{compat=1.15}
\pgfplotsset{every tick label/.append style={font=\small}}

\usepackage[breaklinks]{hyperref}
\usepackage{graphicx}
\newcommand\sbullet[1][.5]{\mathbin{\vcenter{\hbox{\scalebox{#1}{$\bullet$}}}}}
\usepackage{tikz-cd}

\usepackage[ruled]{algorithm}
\usepackage{algorithmic}

\usepackage{gensymb}

\usepackage{placeins}

\usepackage[%
    activate={true,nocompatibility},%
    final,tracking=true,%
    kerning=true,%
    factor=1100,%
    stretch=10,%
shrink=10]{microtype}

\usepackage{tabu}
\usepackage{tabularx}

\microtypecontext{spacing=nonfrench}

\makeatletter 
\let\oldtheequation\theequation
\renewcommand\tagform@[1]{\maketag@@@{\ignorespaces#1\unskip\@@italiccorr}}
\renewcommand\theequation{(\oldtheequation)} 
\makeatother

\begin{document}

\title{Spiking Neural Networks for Nonlinear Regression}

\author{Alexander Henkes, 
        Jason K. Eshraghian,~\IEEEmembership{Member,~IEEE,}
        Henning Wessels
\thanks{Manuscript received October 26, 2022}}

\markboth{Submitted to IEEE Transactions on Neural Networks and Learning Systems, October 26, 2022}%
{Shell \MakeLowercase{\textit{et al.}}: A Sample Article Using IEEEtran.cls for
IEEE Journals}


\maketitle

\begin{abstract}
Spiking neural networks, also often referred to as the third generation of
neural networks, carry the potential for a massive reduction in memory and
energy consumption over traditional, second-generation neural networks.
Inspired by the undisputed efficiency of the human brain, they introduce
temporal and neuronal sparsity, which can be exploited by next-generation
neuromorphic hardware. To broaden the pathway toward engineering applications, where regression tasks are omnipresent, we
introduce this exciting technology in the context of continuum mechanics.
However, the nature of spiking neural networks poses a challenge for regression
problems, which frequently arise in the modeling of engineering sciences. To
overcome this problem, a framework for regression using spiking neural networks
is proposed. In particular, a network topology for decoding binary spike trains
to real numbers is introduced, utilizing the membrane potential of spiking
neurons. Several different spiking neural architectures, ranging from simple
spiking feed-forward to complex spiking long short-term memory neural networks,
are derived. Numerical experiments directed towards regression of linear
and nonlinear, history-dependent material models are carried out. As SNNs exhibit memory-dependent dynamics, they are a natural fit for modelling history-dependent materials which are prevalent through all of engineering sciences. For example, we show that SNNs can accurately model materials that are stressed beyond reversibility, which is a challenging type of non-linearity. A direct
comparison with counterparts of traditional neural networks shows that the
proposed framework is much more efficient while retaining precision and
generalizability. All code has been made publicly available in the interest of
reproducibility and to promote continued enhancement in this new domain.
\end{abstract}

\begin{IEEEkeywords}
        artificial neural networks, spiking neural networks, regression,
        continuum mechanics, neuromorphic hardware
        \end{IEEEkeywords}

\section{Introduction}
\IEEEPARstart{I}{n} recent years, artificial neural networks (ANN) have gained
much attention in the engineering sciences and applied mathematics due to their
flexibility and universal approximation capabilities, both for functions
\cite{hornik1989multilayer, cybenko1989approximation} and operators
\cite{chen1995universal}. Their outstanding but surprising generalizability
capabilities are yet to be understood \cite{berner2021modern}. In computational
engineering sciences, their advantages have been utilized in a variety of
applications, including fluid dynamics \cite{kutz2017deep, zhang2021artificial,
cai2022physics, raissi2020hidden, wessels2020neural}, solid mechanics
\cite{haghighat2021physics, buffa2012mechanical, abueidda2021meshless,
nie2020stress}, micromechanics \cite{henkes2021deep, henkes2021physics,
mianroodi2021teaching, wessels2022computational}, material parameter
identification \cite{dehghani2020poroelastic, thakolkaran2022nn,
zhang2020physics, zhang2022analyses, anton2021identification}, constitutive
modeling \cite{as2022mechanics, yang2020exploring, xu2021learning,
fernandez2020application}, fracture mechanics \cite{aldakheel2021feed,
liu2002detection}, microstructure generation \cite{henkes2022three,
hsu2021microstructure, mosser2017reconstruction}, contact problems
\cite{ma2021data, oner2022plane, ardestani2014feed, feng2022prediction} heat
transfer \cite{cai2021physics, laubscher2021simulation, tamaddon2020data,
niaki2021physics} and uncertainty quantification \cite{felipe2020machine,
balokas2018neural, fuhg2022interval, olivier2021bayesian}, among numerous
others. See \cite{bock2019review, kumar2021machine, blechschmidt2021three} for
review publications.

Despite the success of ANNs, several problems arise alongside their utilization,
such as the need for high-frequency memory access, which leads to high
computational power demand \cite{roy2019towards, indiveri2015memory}. This results in huge costs for training and often
makes it preferable to run inference in remote servers during deployment. In
general, ANNs are most often trained on GPUs, whose energy consumption is
problematic in embedded systems (e.g., sensor devices) as is required in
automotive and aerospace applications \cite{burr2017neuromorphic}. Furthermore, high latency during
prediction time can arise where acceleration or parallelization is not
available.

Originally motivated by the human brain, today's traditional ANN architectures
are an oversimplification of biology, relying on dense matrix multiplication.
From a numerical and computational hardware point of view, dense matrix
multiplication is often suboptimal. Sparsity is thought to be favorable as it reduces
dependence on memory access and data communication \cite{perez2021sparse}.
In contrast, the human brain is much more efficient, where neurons are
considered to be sparsely activated \cite{olshausen2006other}. This stems from
the fact that the brain uses sparse electronic signals for information
transmission instead of dense activations. This leads to remarkable capabilities
by using only about 10-20 watts of energy. One attempt to overcome these
drawbacks of ANNs is to introduce the information transmission mechanism of
biological neurons into network architectures. These networks are called
\textit{spiking neural networks} (SNN) due to the electronic impulses or
\textit{spikes} used for communication between neurons
\cite{gerstner2002spiking}. This leads to sparse activations, which can be
efficiently exploited by \textit{neuromorphic} hardware, such as Loihi
\cite{davies2018loihi}, SpinNaker \cite{furber2014spinnaker}, and TrueNorth
\cite{merolla2014million}.  It has been shown that these specialized hardware
chips are able to reduce the energy consumption of neural network-based
processes by factors of up to $\times$1000 \cite{davies2018loihi,
azghadi2020hardware, frenkel2022reckon, ceolini2020hand, orchard2021efficient}.

What was classically in the domain of neuroscientists recently has been
investigated in the context of deep learning, e.g., the adoption of SNNs to
supervised learning as popularised with traditional ANNs in frameworks such as
TensorFlow \cite{tensorflow2015-whitepaper} and PyTorch \cite{NEURIPS2019_9015},
resulting in similar frameworks for spiking deep learning like snnTorch
\cite{eshraghian2021training}. Some applications of spiking deep learning
includes image processing using a spiking ResNet \cite{fang2021deep} and
temporal data processing using spiking LSTM variants \cite{bellec2018long,
rao2022long}. A combination of spiking convolutional neural networks and LSTMs
was proposed in \cite{yang2022neuromorphic}. SNNs have been used for image
segmentation \cite{patel2021spiking} and localization \cite{barchid2022spiking,
moro2022neuromorphic}.

To the best of the author's knowledge, the scope of regression modeling using
SNNs remains limited. In \cite{iannella2001spiking}, an architecture using
\textit{inter-spike interval temporal encoding} has been proposed, where learned
functions were limited to piecewise constant functions. In
\cite{gehrig2020event}, a SNN was used for the regression of angular velocities
of a rotating event camera.  Building on these results,
\cite{ranccon2021stereospike} proposed a SNN for depth reconstruction. In
\cite{kahana2022function}, a DeepONet \cite{lu2021learning} using SNN was
proposed, which used a floating point decoding scheme to regress on simple
one-dimensional functions. In \cite{shrestha2018slayer}, gradient descent was
applied to learn spike times, and in \cite{eshraghian2022memristor}
classification problems were recast as regression tasks in the context of
memristor based hardware. The focus of the present work lays on neuromorphic
hardware, which is specifically designed for SNNs.

As regression problems are omnipresent in engineering sciences, a flexible and
broadly applicable framework would enable SNNs to be utilized in a variety of
engineering applications and further unfold the potential of neuromorphic
hardware. To this end, the present study aims towards the following key
contributions: 

\begin{itemize} 
    \item \textbf{Introduction of spiking neural networks:} Concise introduction
        of this emerging technique. Open source benchmark code for the research
        community.
    \item \textbf{History dependent regression framework:} SNNs are naturally
        suited for classification. Engineering problems often involve regression
        tasks. We present a flexible framework to use SNNs in complex regression
        tasks, namely history-dependent material behavior in the case of
        isotropic hardening plasticity. As such, we demonstrate that SNNs can
        model systems that exhibit hysteresis.
    \item \textbf{Efficiency, sparsity and latency:} We benchmark our SNN on
        neuromorphic hardware in terms of energy consumption as compared to
        non-spiking equivalent networks, demonstrating that they are much more
        efficient with respect to memory and power consumption, making neural
        networks more sustainable. Their deployment on neuromorphic hardware
        allows highly efficient usage in embedded environments. We present a
        detailed comparison with standard ANN for memory consumption and power
        consumption. 
\end{itemize}

\noindent
The present work intends to introduce this important novel technique to the
community of computational mechanics and applied mathematics. To concentrate on
the novelties and keep the presentation concise, we restrict ourselves to
one-dimensional, history-dependent regression problems. However, the framework
is not restricted to single-variable regression and is easily applicable to a
multivariable regression.  Furthermore, we explicitly do not consider advanced
modeling concepts that ensure the thermodynamical consistency of the material
models at hand. Nevertheless, our framework can be easily extended towards these
important constraints by utilizing works from, e.g., \cite{kalina2022automated,
masi2021thermodynamics, masi2022multiscale}. The latter are translatable from
classical ANNs to SNNs.

The remainder of this paper is structured as follows. In \autoref{sec:mysnn},
the basic notations of SNNs are derived from traditional ANNs. A simple spiking
counterpart to the classical densely connected feed-forward neural network is
introduced. After that, our regression SNN topology is proposed. First
applications toward linear elasticity point out the problems arising in SNN
regression. This basic architecture is extended towards recurrent feedback loops
in \autoref{sec:example2}. The ability of these recurrent SNNs is showcased on a
nonlinear material model. To process history-dependent regression tasks with
dependencies over a large number of time steps, a spiking LSTM is introduced in
\autoref{sec:example3}. An application to a history-dependent plasticity model
shows that SNNs can achieve similar accuracies as their traditional counterparts
while being much more efficient. The paper closes with a conclusion and an
outlook toward future research directions in \autoref{sec:conclusion}. For the
code accompanying this manuscript, see the \textit{data availability} section at
the end of this manuscript.

\section{SNN for regression}
\label{sec:mysnn} \noindent
SNNs are considered to be the third generation of neural networks. While the first generation
was restricted to shallow networks, the second generation is characterized by
deep architectures. A broad use of 2$^{nd}$ generation neural networks has been
enabled by the availability of automatic differentiation and software frameworks
such as Tensorflow \cite{tensorflow2015-whitepaper}. To introduce spiking neural
networks, we compare them with their well-known 2$^{nd}$ generation
counterparts. Our notation follows \cite{eshraghian2021training}. Standard works
in theoretical neuroscience include \cite{dayan2005theoretical,
izhikevich2007dynamical} and \cite{gerstner2014neuronal}. Several overviews of
SNNs with respect to deep learning can be found in \cite{tavanaei2019deep,
pfeiffer2018deep, richards2019deep}. First, the standard feed-forward densely
connected ANN is introduced. After that, a basic SNN is derived from this.

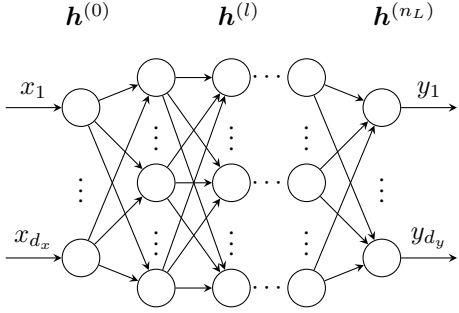
\begin{figure}[htb] 
    \centering 
    \tikzset{%
        every neuron/.style={ circle,
        draw, minimum size=0.5cm }, neuron missing/.style={ draw=none, scale=1,
    execute at begin node=\color{black}$\vdots$ }, } 
    \begin{tikzpicture}[x=1cm,y=1cm, >=stealth] 
        \node at (0.1,-0.75) [] {$\bm{h}^{(0)}$}; \node at (2.1,-0.75) []
        {$\bm{h}^{(l)}$}; \node at (4.3,-0.75) [] {$\bm{h}^{(n_L)}$}; 

        \foreach \m/\l [count=\y] in {1, missing, 2} \node [every
        neuron/.try, neuron \m/.try] (input-\m) at (0,-1-\y) {};

        \foreach \m [count=\y] in {1, missing, 2, missing,3} \node [every
        neuron/.try, neuron \m/.try ] (hidden-\m) at (1,-0.9-\y*0.7) {};

        \foreach \m [count=\y] in {1, missing, 2, missing,3} \node [every
        neuron/.try, neuron \m/.try ] (hidden2-\m) at (2,-0.9-\y*0.7) {};

        \foreach \m [count=\y] in {1, missing, 2, missing,3} \node [every
        neuron/.try, neuron \m/.try ] (hidden3-\m) at (3,-0.9-\y*0.7) {};

        \foreach \m [count=\y] in {1, missing, 2} \node [every neuron/.try,
        neuron \m/.try ] (output-\m) at (4,-1-\y) {};

        \draw [<-] (input-1) -- ++(-1,0) node [above, midway] {$x_1$}; \draw
        [<-] (input-2) -- ++(-1,0) node [above, midway] {$x_{d_x}$};

        \draw [->] (output-1) -- ++(1,0) node [above, midway] {$y_1$}; \draw
        [->] (output-2) -- ++(1,0) node [above, midway] {$y_{d_y}$};

        \foreach \i in {1,...,2} \foreach \j in {1,...,3} \draw [->]
        (input-\i) -- (hidden-\j);

        \foreach \i in {1,...,3} \foreach \j in {1,...,3} \draw [->]
        (hidden-\i) -- (hidden2-\j);

        \node at (2.5,-1.6) [] {$\cdots$}; \node at (2.5,-3.0) []
        {$\cdots$}; \node at (2.5,-4.4) [] {$\cdots$};

        \foreach \i in {1,...,3} \foreach \j in {1,...,2} \draw [->]
    (hidden3-\i) -- (output-\j); \end{tikzpicture} 
    \caption{Densely connected feed forward neural network topology of an ANN
    $\mathcal{N}(\bm{x}; \bm{\theta})$ as described in \autoref{eq:ann}.} 
    \label{fig:ann} 
\end{figure}

An ANN is a parametrized, nonlinear function composition. The \textit{universal
function approximation theorem} \cite{hornik1989multilayer} states that
arbitrary Borel measurable functions can be approximated with ANNs. There are
several different architectures for ANNs, e.g., feed-forward, recurrent, or
convolutional networks, which can be found in standard references such as
\cite{bishop2006pattern, goodfellow2016deep, aggarwal2018neural, geron2019hands,
chollet2018deep}. Following \cite{hauser2018principles}, most ANN formulations
can be unified. An ANN $\mathcal{N}$, more precisely, a \textit{densely
connected feed-forward neural network}, is a function from an \textit{input
space} $\mathbb{R}^{d_x}$ to an \textit{output space} $\mathbb{R}^{d_y}$,
defined by a composition of nonlinear functions $\bm{h}^{(l)}$, such that 
\begin{align} 
    \label{eq:ann} 
    \mathcal{N}: \mathbb{R}^{d_x} 
    &\to \mathbb{R}^{d_y} \\ 
    \bm{x} 
    &\mapsto \mathcal{N}(\bm{x}) = \bm{h}^{(l)} \circ \ldots \circ \bm{h}^{(0)} = \bm{y}, \nonumber \\ l &= 1, \ldots, n_L. \nonumber
\end{align} 
Here, $\bm{x}$ denotes an \textit{input vector} of dimension $d_x$ and $\bm{y}$
an \textit{output vector} of dimension $d_y$. The nonlinear functions
$\bm{h}^{(l)}$ are called \textit{layers} and define an $l$-fold composition,
mapping input vectors to output vectors.  Consequently, the first layer
$\bm{h}^{(0)}$ is defined as the \textit{input layer} and the last layer
$\bm{h}^{(n_L)}$ as the \textit{output layer}, such that 
\begin{equation}
    \bm{h}^{(0)} = \bm{x} \in \mathbb{R}^{d_x}, \qquad \bm{h}^{(n_L)} = \bm{y}
    \in \mathbb{R}^{d_y}. 
    \label{eq:layer} 
\end{equation} 
The layers $\bm{h}^{(l)}$ between the input and output layer, called
\textit{hidden layers}, are defined as
\begin{align} 
    \label{eq:hidden} 
    \bm{h}^{(l)} &= \left\{h_{\eta}^{(l)}, \; \eta = 1,
        \ldots,
    n_{u}\right\}, \\ h_{\eta}^{(l)} &= 
    \phi^{(l)}\left(\bm{W}^{(l)}_{\eta} \bm{h}^{(l-1)}\right),
    \nonumber
\end{align} 
where $h_{\eta}^{(l)}$ is the $\eta$-th \textit{neural unit} of the $l$-th layer
$\bm{h}^{(l)}$, $n_u$ denotes the \textit{total number of neural units per
layer}, $\bm{W}_{\eta}^{(l)}$ is the \textit{weight vector} of the $\eta$-th
neural unit in the $l$-th layer $\bm{h}^{(l)}$ and $\bm{h}_{}^{(l-1)}$ is
the output of the preceding layer, where bias terms are absorbed
\cite{aggarwal2018neural}. Furthermore, $\phi^{(l)}: \mathbb{R} \to \mathbb{R}$
is a nonlinear \textit{activation function}. All weight vectors
$\bm{W}_{\eta}^{(l)}$ of all layers $\bm{h}^{(l)}$ can be gathered in a single
expression, such that 
\begin{equation}
    \bm{\theta}_{\text{ANN}}=\left\{\bm{W}_{\eta}^{(l)}\right\}, 
    \label{eq:parameters}
\end{equation} 
where $\bm{\theta}$ inherits all parameters of the ANN $\mathcal{N}(\bm{x})$
from \autoref{eq:ann}. Consequently, the notation $\mathcal{N}(\bm{x};
\bm{\theta})$ emphasizes the dependency of the outcome of an ANN on the input on
the one hand and the current realization of the weights on the other hand. The
specific combination of layers $\bm{h}^{(l)}$ from \autoref{eq:hidden}, neural
units $h_{\eta}^{(l)}$ and activation functions $\phi^{(l)}$ from
\autoref{eq:hidden} is called \textit{topology} of the ANN $\mathcal{N}(\bm{x};
\bm{\theta})$. The weights $\bm{\theta}$ from \autoref{eq:parameters} are
typically found by gradient-based optimization with respect to a task-specific
$\textit{loss function}$ \cite{goodfellow2016deep}. An illustration of a densely
connected feed-forward ANN is shown in \autoref{fig:ann}.

It can be seen that the ANN described in \autoref{eq:ann} takes an input
$\bm{x}$ and produces an output $\bm{y}$, one at a time. If history-dependent
input and output data $\bm{x}_t \in \mathbb{R}^{d_t \times d_x}$ and $\bm{y}_t
\in \mathbb{R}^{d_t \times d_y}$ is considered, the formulation of the hidden
layers reads
\begin{align} 
    \label{eq:hidden_t} 
    \bm{h}_t^{(l)} &= \left\{h_{\eta, t}^{(l)}, 
        \; \eta = 1, \ldots, n_{u},
        \; t = 0, \ldots, d_{t}
    \right\}, \\ h_{\eta, t}^{(l)} &= 
    \phi^{(l)}\left(\bm{W}^{(l)}_{\eta} \bm{h}_t^{(l-1)}\right),
    \nonumber
\end{align} 
where the time component is discrete. This can be understood as processing each
discrete-time slice of the input vector of the preceding layer
$\bm{h}_{t=0}^{(l-1)} \rightarrow \bm{h}_{t=1}^{(l-1)} \rightarrow, \ldots,
\rightarrow \bm{h}_{t=d_t}^{(l-1)} $ sequentially, where the weights
$\bm{W}^{(l)}_{\eta}$ are shared over all time steps. At this stage, the
formulation in \autoref{eq:hidden_t} is purely notationally, as there is no
connection of the weights through different time steps. 

Now, a SNN can be seen as a history-dependent ANN,
which introduces memory effects by means of
biologically inspired processes. To this end, the activation function
$\phi^{(l)}$ in \autoref{eq:hidden_t} can be formulated as
\begin{align}
    \phi_{\text{spk}, t}^{(l)} = 
    \begin{cases} 
        1, \quad U_{\eta, t}^{(l)} \geq U_{\text{thr}, \eta}^{(l)} \\ 
        0, \quad U_{\eta, t}^{(l)} < U_{\text{thr}, \eta}^{(l)},
    \end{cases} 
    \label{eq:snn} 
\end{align}
with
\begin{equation}
    U_{\eta, t}^{(l)} = \beta_{\eta}^{(l)} U_{\eta, t - 1}^{(l)} +
    \bm{W}^{(l)}_{\eta} \bm{h}_{t}^{(l-1)} - \phi_{\text{spk}, t - 1}^{(l)}
    U_{\text{thr}, \eta}^{(l)},
    \label{eq:membrane_potential}
\end{equation}
where $U_{\eta, t}^{(l)}$ is the \textit{membrane potential} of the $\eta$-th
neural unit at time $t$, $U_{\text{thr}, \eta}^{(l)}$ denotes the
\textit{membrane threshold}, $\beta_\eta^{(l)}$ is the \textit{membrane
potential decay rate} and $\bm{W}^{(l)}_{\eta} \bm{h}_{t}^{(l-1)}$ is the
standard ANN weight multiplied with the preceding layer of the current time
step, respectively, see \autoref{eq:hidden_t}. Basically, the SNN activation
restricts the neural unit to output discrete pulses $(\phi_{\text{spk}}=1)$ if
the membrane threshold is reached by the time-evolving membrane potential, or to
remain silent $(\phi_{\text{spk}}=0)$. These pulses are called \textit{spikes}.
The last summand in \autoref{eq:snn}, $- \phi_{\text{spk}, t - 1}^{(l)}
U_{\text{thr}, \eta}^{(l)}$, is called the \textit{reset mechanism} and resets the
membrane potential by the threshold potential once a spike is emitted. The
membrane threshold and membrane potential decay rate can be optimized during
training, such that the optimization parameters of a SNN are 
\begin{equation}
    \bm{\theta}_{\text{SNN}}=\left\{\bm{W}_{\eta}^{(l)}, \beta_{\eta}^{(l)},
    U_{\text{thr}, \eta}^{(l)}\right\}.
    \label{eq:parameters_SNN}
\end{equation} 
The SNN formulation in \autoref{eq:membrane_potential} is called the \textit{leaky
integrate and fire} (LIF) neuron model, and is one of the most widely used models in spike-based
deep learning. It can be seen as the baseline SNN and plays a similar role as
densely connected feed-forward ANN in classical deep learning.

The formulation in \autoref{eq:membrane_potential} can be seen as the explicit
forward Euler solution of an ordinary differential equation, describing the time
variation of the membrane potential, see \cite{eshraghian2021training} for
details. 

\begin{figure}[htb] 
    \centering 
    \includegraphics[width=0.3\textwidth]{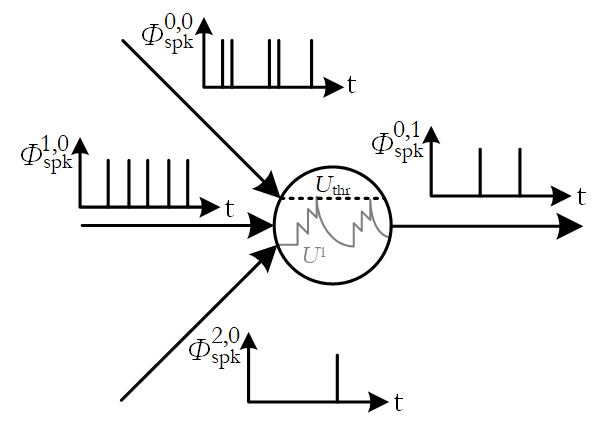}
    \caption{Spiking Neuron Dynamics. Input spikes $\phi^{i,0}_{\rm spk}$
    trigger changes in the membrane potential $U^1$, which when sufficiently
excited beyond a threshold $U_{\rm thr}$ causes the neuron to emit an output
spike $\phi^{j,1}_{\rm spk}$.} 
    \label{fig:snntorch_1} 
\end{figure}

The main difference between SNNs and classical ANNs lies in the way information
is processed and propagated through the network from neuron to neuron. In
standard ANNs, inputs, hidden layers, and output vectors are handled via dense
matrices. In spiking neural networks, sparsity is introduced by utilizing
spikes, which are single events expressed via a Dirac delta function or a
discrete pulse in continuous or discrete settings, respectively. A group of
spikes over time is called a \textit{spike train} $\bm{i} = \left[ i_t, \quad t
= 0, ..., n_t \right]$. To this end, a spiking neuron is subjected to a spike
train over a time interval, consisting of spikes $(1)$ or zero input $(0)$. The
membrane potential $U_{\eta, t}^{(l)}$ is modulated with incoming spikes $i_t$. In the
absence of input spikes, the membrane voltage decays over time due to the
membrane decay rate $\beta_{\eta}^{(l)}$. The absence of spikes introduces
sparsity because in every time step, the neural unit output is constrained to
either zero or one. This fact can be exploited on \textit{neuromorphic
hardware}, where memory and synaptic weights need only be accessed if a spike is
apparent. Otherwise, no information is transmitted. In contrast, conventional
ANNs do not leverage sparsely activated neurons, and most deep learning
accelerators, such as GPUs or TPUs, are correspondingly not optimized for it.

\begin{figure}[htb]
     \centering
         \includegraphics[width=0.3\textwidth]{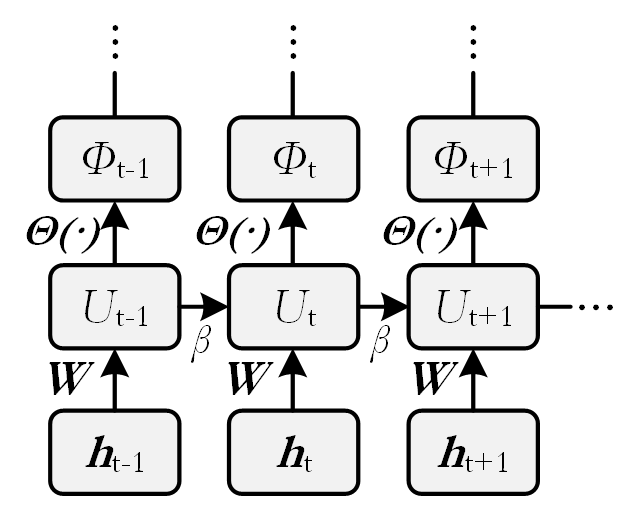}
        \caption{Computational graph of an unrolled SNN. Forward-pass.}
        \label{fig:three_graphs_forward}
\end{figure}

\begin{figure}[htb]
     \centering
     \includegraphics[width=0.3\textwidth]{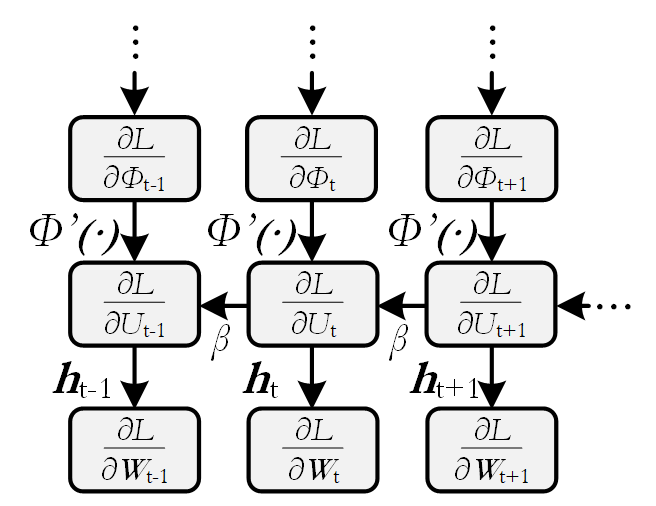}
        \caption{Computational graph of an unrolled SNN. Backward-pass.}
        \label{fig:three_graphs_back}
\end{figure}

Unfortunately, the spiking activation $\phi_{\text{spk}, t}^{(l)}$ in
\autoref{eq:snn} is non-differentiable. To use the backpropagation algorithm
from standard ANNs, the activation is replaced using a \textit{surrogate
gradient} during the backward pass. Several different formulations have been
proposed, see, e.g., \cite{neftci2019surrogate, perez2021sparse,
zenke2018superspike}. In this work, the \textit{arcus tangent surrogate
activation} from
\cite{fang2021incorporating} is used:
\begin{equation}
    \phi_{\text{surr}} \left( x \right) = \frac{1}{\pi} \operatorname{arctan}
    \left( \pi x \right), \qquad \phi_{\text{surr}}' \left( x \right) =
    \frac{1}{1 + \left( \pi x \right)^2 }, 
\end{equation}
for some input $x$. The surrogate $\phi_{\text{surr}} \left( x \right)$ is
continuously differentiable and preserves the gradient dynamics of the network.
Thus, for training using backpropagation and its variants, $\phi_{\text{surr}}'$
is employed. Illustrations can be found in \autoref{fig:three_graphs_forward} and \autoref{fig:three_graphs_back}.

\subsection{Network topology}
\label{sec:topology} 
\begin{figure}[htb] 
    \centering 
    \includegraphics[width=0.45\textwidth]{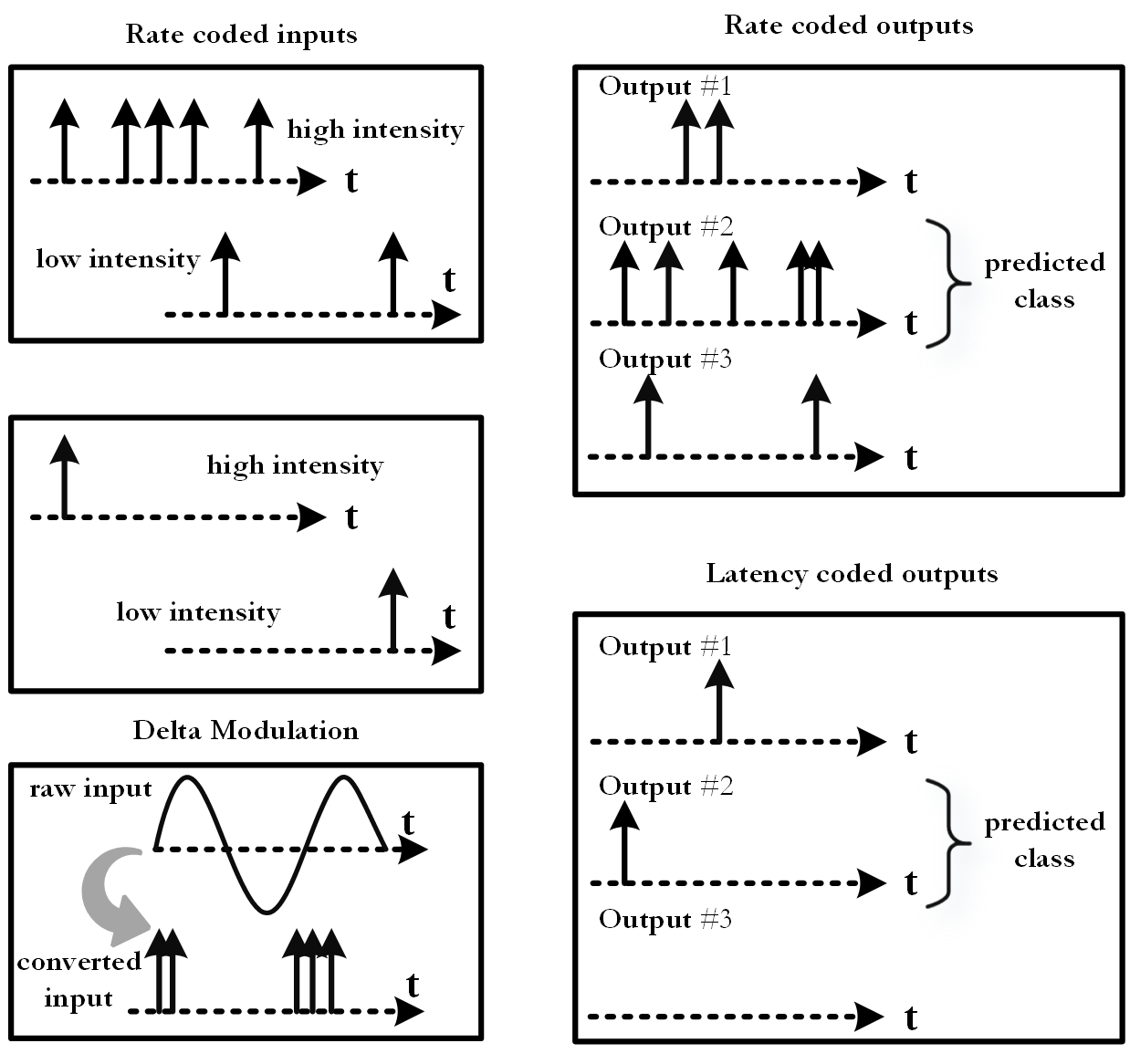}
    \caption{A sample of spike-based encoding and decoding strategies. Left:
    Real-valued inputs are encoded into spikes by means of different strategies
\cite{eshraghian2021training}, e.g., (1)  high intensities or large values
result in a large number of spikes (top left), (2) high intensities or large
values result in early spike firing (center left), (3) delta modulation where
spikes are produced for positive gradients of the input function (bottom left).
Right: In classification, the predicted class is determined via (1) the number
of spikes (top right) or (2) the first occurence of spikes (bottom right).
Regression is introduced in section \autoref{sec:linear_elasticity}. }
    \label{fig:coding} 
\end{figure}

\noindent
The key question for using SNN in regression is how to transform real input
values into binary spikes and binary spike information at the output layer back
to real numbers. The former task is called \textit{spike encoding}, whereas the
latter is called \textit{spike decoding}. In this work, a \textit{constant
current injection} is chosen for the encoding part, whereas a novel
\textit{population voting on membrane potential} approach is chosen for the
decoding part. Other forms of encoding include \textit{rate encoding},
\textit{latency encoding} and \textit{delta modulation}, among others.
Similarly, different decoding strategies exist, such as \textit{rate decoding}
and \textit{latency decoding}. An illustration of various encoding and decoding
strategies is shown in \autoref{fig:coding}. See \cite{eshraghian2021training}
for an overview and detailed description. 

All network topologies used in the upcoming numerical examples follow a general
scheme, which is flexible and suited for regression tasks. First, the real input
$\bm{x}_t$ is provided as a constant input to the first layer $\bm{h}_t^{(0)}$,
for all time steps $t$, such that
\begin{equation}
    \bm{h}_t^{(0)} \left( \bm{x}_t \right) = \bm{h}_t^{\text{const}} \left(
    \bm{x}_t \right) = \bm{x}_t \; \forall \; t \in \left[ 0, d_t \right].
\end{equation}
Then, several SNN layers $\bm{h}_t^{(l)}$ follow, where the exact formulation is
arbitrary, and will be given for every numerical example. The output of the last
spiking layer $\bm{h}_t^{(n_L)}$ is transformed into a \textit{decoding layer}
$\bm{h}_t^{\text{dec}}$, which takes the membrane potential of every time step
as input and outputs real numbers
\begin{equation}
    \bm{h}_t^{\text{dec}} = \beta_{\eta}^{(l)} U_{\eta, t - 1}^{(l)} +
    \bm{W}^{(l)}_{\eta} \bm{h}_{t}^{(l-1)},
    \label{eq:dec}
\end{equation}
which is essentially the formulation of \autoref{eq:membrane_potential}, where
no spikes and reset mechanisms are used. The transformed values are then
transferred to the `\textit{population voting layer}', where the output of all
neurons of the decoding layer are averaged to give real numbers. This results in
\begin{equation}
\bm{h}_t^{\text{pop}} = \frac{1}{n_o} \sum_{n_o} \left( \beta_{\eta}^{(l)}
U_{\eta, t - 1}^{(l)} + \bm{W}^{(l)}_{\eta} \bm{h}_{t}^{(l-1)} \right),
    \label{eq:pop}
\end{equation}
where $n_o$ denotes the number of neurons in the population voting layer and
again, no spikes or reset mechanisms are used.

\begin{figure}[htb]
    \centering
    \begin{tikzpicture}[]
        \node (x) at (1.5, 1.0) 
        {$\bm{x}_t$};

        \node (const) at (1.5, 0.0) [rectangle, draw, thick] 
        {$\bm{h}_t^{\text{const}}$};

        \draw[->, thick] (x) to (const);

        \node (snn1) at (3.0, 0.0) [rectangle, draw, thick] 
        {$\bm{h}_t^{(1)}$};

        \draw[->, thick] (const) to (snn1);

        \node (snnl) at (4.5, 0.0) [rectangle, draw, thick] 
        {$\bm{h}_t^{(l)}$};

        \node[draw=none] at (3.8, 0.0) {$\dots$};

        \node (snnnL) at (6.0, 0.0) [rectangle, draw, thick] 
        {$\bm{h}_t^{(n_L)}$};

        \node[draw=none] at (5.2, 0.0) {$\dots$};

        \node (dec) at (7.5, 0.0) [rectangle, draw, thick] 
        {$\bm{h}_t^{\text{dec}}$};

        \draw[->, thick] (snnnL) to (dec);

        \node (pop) at (9.0, 0.0) [rectangle, draw, thick] 
        {$\bm{h}_t^{\text{pop}}$};

        \draw[->, thick] (dec) to (pop);

        \node (y) at (9.0, -1.0) 
        {$\bm{y}_t$};

        \draw[->, thick] (pop) to (y);
    \end{tikzpicture}
    \caption{Topology of the spiking regression network introduced in
    \autoref{sec:topology}.}
    \label{fig:topology}
\end{figure}
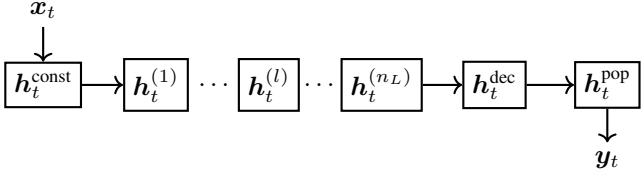

The final spiking regression topology network $\mathcal{S}$ can be written as
\begin{align} 
    \mathcal{S}_{}:
    \mathbb{R}^{d_t \times d_x} &\to \mathbb{R}^{d_t \times d_y} &\\
    \bm{x}_t &\mapsto  
    \bm{h}_t^{\text{pop}} \circ
    \bm{h}_t^{\text{dec}} \circ
    \bm{h}_t^{(n_L)} \circ \ldots \circ
    \bm{h}_t^{(l)} \circ \ldots & \nonumber \\
    &\ldots \circ \bm{h}_t^{(1)} \circ
    \bm{h}_t^{\text{const}} \left( \bm{x}_t \right) = \bm{y}_t. &  \nonumber
    \label{eq:snn_topology} 
\end{align} 
To summarize, information flows in the form of constant current (real numbers)
into the input layer $\bm{h}_t^{\text{const}}$, is then transformed into binary
spikes in the spiking layers $\bm{h}_t^{(l)}$ and transformed back into
real numbers in the translation layer $\bm{h}_t^{\text{dec}}$. The output is
formed in the population layer $\bm{h}_t^{\text{pop}}$. A graphical
interpretation is given in \autoref{fig:topology}.

For all the following numerical examples, the AdamW optimizer from
\cite{loshchilov2017decoupled} is used. The parameter are set as follows:
learning rate $\alpha = 1 \times 10^{-3}$, exponential decay rates for the first
and second moment estimates $\beta_1 = 0.9$ and $\beta_2 = 0.999$, respectively,
weight decay $\lambda = 0.01$. The training was carried out on a Nvidia GeForce
RTX 3090 GPU using snnTorch \cite{eshraghian2021training} and PyTorch
\cite{NEURIPS2019_9015}. In this work, the mean relative error $\mathcal{E}$ is
used, which is defined as
\begin{equation}
    \mathcal{E}_{\bullet}(\bullet) = 
    \frac{1}{n_s}
    \displaystyle\sum_{i=1}^{n_s}
    \frac{||\hat{{\bullet}}_i - {\bullet}_i||_2}{||\hat{{\bullet}}_i||_2},
    \label{eq:error}
\end{equation}
for some input $\bullet$ and baseline $\hat{\bullet}$. If the error is reported
for all time steps, $\bullet$ is a vector containing the values of all time
steps. If the error is reported for the last time step, $\bullet$ is equal to
the last component of the corresponding history-dependent vector.

\subsection{Numerical experiment: Linear elasticity}
\label{sec:linear_elasticity}
\noindent
The first study investigates the effect of the number of time steps on the
performance of the proposed LIF topology in a simple linear regression problem.
To this end, the general model described in \autoref{sec:topology} with LIF
defined in \autoref{eq:membrane_potential} is used, resulting in the following
network topology
\begin{equation} 
    \mathcal{S} \left( \varepsilon_t \right) = 
    \bm{h}_t^{\text{pop}} \circ
    \bm{h}_t^{\text{dec}} \circ
    \bm{h}_t^{\text{LIF}} \circ 
    \bm{h}_t^{\text{LIF}} \circ 
    \bm{h}_t^{\text{LIF}} \circ 
    \bm{h}_t^{\text{const}} \left( \varepsilon_t \right) = \sigma_t.  
    \label{eq:LIF} 
\end{equation} 

To begin with, a simple linear elastic material model with strains in the range
of $\varepsilon = [0, 0.001]$ and fixed Young's modulus $E = 2.1 \times 10^5$
MPa is considered, such that the resulting stress $\sigma$ is
\begin{equation}
    \sigma = E \varepsilon.
    \label{eq:elasticity}
\end{equation}

The training data consists of strain input, uniformly sampled in the interval
$\varepsilon = [0, 0.001]$, and stress output calculated according to
\autoref{eq:elasticity}. Three datasets are generated, namely a training set, a
validation and a test set consisting each of $n_{\text{train}} = n_{\text{val}}
= n_{\text{test}} = 1024$ samples. All three datasets are standardized using the
mean and standard deviation from the training set. The batch size is chosen as
$n_{\text{batch}} = 1024$. The number of neurons $n_u$ is chosen as $n_u = 128$
and is kept constant over all layers. The training is carried out for $2 \times
10^3$ epochs. The model performing best on the validation set is chosen for
subsequent evaluations. The mean relative error accumulated over all time steps
and the mean relative error of the last time step with respect to the test set
are reported.

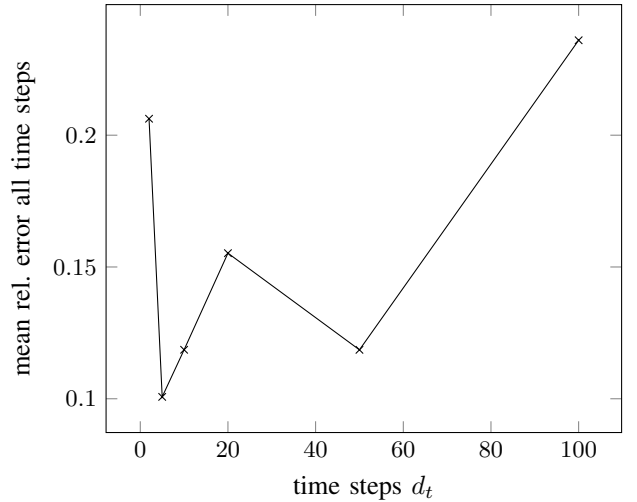
\begin{figure}[htb]
    \centering
    \begin{tikzpicture}
        \begin{axis}[ 
                xlabel={time steps $d_t$},
                ylabel={mean rel. error all time steps},
            ] 

            \addplot[color=black, mark=x, mark options=solid] 
            coordinates {
                (2, 0.20624607801437378)
                (5, 0.10071968287229538)
                (10, 0.11858871579170227)
                (20,0.15524239838123322)
                (50, 0.1185651645064354)
                (100, 0.2360144555568695)
            };
        \end{axis}
    \end{tikzpicture}
    \caption{\textbf{Elasticity - error of all timesteps:} Mean relative error
    for all time steps with respect to the different total number of time steps.
The error is rising for a larger number of time steps.}
    \label{fig:elasticity_timesteps} 
\end{figure}

\begin{figure}[htb]
    \centering
        \begin{tikzpicture}[]
            \begin{axis}[ 
                    xlabel={strain},
                    ylabel={stress in [MPa]},
                    legend pos=south east
                ] 
                \addplot[color=black] 
            coordinates {
                                (0.0, 0.0)
                                (3.5886652767658234e-05, 7.53619384765625)
                                (7.177330553531647e-05, 15.0723876953125)
                                (0.00010765972547233105, 22.60858154296875)
                                (0.00014354661107063293, 30.144744873046875)
                                (0.00017943303100764751, 37.68096923828125)
                                (0.0002153199166059494, 45.217071533203125)
                                (0.000251206336542964, 52.7532958984375)
                                (0.00028709275647997856, 60.28948974609375)
                                (0.0003229794092476368, 67.82568359375)
                                (0.00035886606201529503, 75.36187744140625)
                                (0.00039475271478295326, 82.89804077148438)
                                (0.00043063913471996784, 90.43423461914062)
                                (0.0004665257874876261, 97.97042846679688)
                                (0.0005024124402552843, 105.50662231445312)
                                (0.0005382990930229425, 113.04278564453125)
                                (0.0005741857457906008, 120.5789794921875)
                                (0.0006100721657276154, 128.11517333984375)
                                (0.0006459588184952736, 135.6513671875)
                                (0.0006818454712629318, 143.18753051757812)
                                (0.0007177320076152682, 150.72372436523438)
                                (0.0007536186603829265, 158.25991821289062)
                                (0.0007895051967352629, 165.79608154296875)
                                (0.0008253918495029211, 173.332275390625)
                                (0.0008612785022705793, 180.86846923828125)
                                (0.0008971650386229157, 188.4046630859375)
                                (0.0009330515749752522, 195.94085693359375)
                                (0.0009689382277429104, 203.47702026367188)
                                (0.0010048248805105686, 211.01324462890625)
                                (0.001040711416862905, 218.54937744140625)
                                (0.0010765980696305633, 226.08560180664062)
                                (0.0011124846059828997, 233.62176513671875)
                                (0.001148371258750558, 241.157958984375)
                                (0.0011842579115182161, 248.69415283203125)
                                (0.0012201444478705525, 256.2303466796875)
                                (0.0012560311006382108, 263.76654052734375)
                                (0.001291917753405869, 271.302734375)
                                (0.0013278044061735272, 278.8388671875)
                                (0.0013636908261105418, 286.37506103515625)
                                (0.0013995774788782, 293.9112854003906)
                                (0.0014354640152305365, 301.44744873046875)
                                (0.0014713506679981947, 308.983642578125)
                                (0.001507237204350531, 316.5198059082031)
                                (0.0015431238571181893, 324.0559997558594)
                                (0.0015790103934705257, 331.5921936035156)
                                (0.001614897046238184, 339.1283874511719)
                                (0.0016507835825905204, 346.66455078125)
                                (0.0016866702353581786, 354.20074462890625)
                                (0.0017225568881258368, 361.7369384765625)
                                (0.001758443540893495, 369.27313232421875)
                                (0.0017943300772458315, 376.809326171875)
                                (0.0018302167300134897, 384.34552001953125)
                                (0.0018661032663658261, 391.8816833496094)
                                (0.0019019899191334844, 399.4178771972656)
                                (0.0019378764554858208, 406.95404052734375)
                                (0.001973763108253479, 414.4902648925781)
                                (0.0020096495281904936, 422.0263977050781)
                                (0.002045536180958152, 429.5625915527344)
                                (0.00208142283372581, 437.0987854003906)
                                (0.0021173092536628246, 444.63494873046875)
                                (0.002153195906430483, 452.171142578125)
                                (0.002189082559198141, 459.70733642578125)
                                (0.0022249692119657993, 467.2435302734375)
                                (0.0022608558647334576, 474.77972412109375)
                                (0.002296742517501116, 482.31591796875)
                                (0.002332629170268774, 489.85211181640625)
                                (0.0023685158230364323, 497.3883361816406)
                                (0.002404402242973447, 504.9244689941406)
                                (0.002440288895741105, 512.460693359375)
                                (0.0024761755485087633, 519.9968872070312)
                                (0.0025120622012764215, 527.5330810546875)
                                (0.002547948621213436, 535.0692138671875)
                                (0.0025838352739810944, 542.6054077148438)
                                (0.0026197219267487526, 550.1416015625)
                                (0.002655608579516411, 557.6777954101562)
                                (0.0026914949994534254, 565.2139282226562)
                                (0.0027273816522210836, 572.7501220703125)
                                (0.002763268304988742, 580.2863159179688)
                                (0.0027991549577564, 587.8225708007812)
                                (0.0028350413776934147, 595.3587036132812)
                                (0.002870928030461073, 602.8948974609375)
                                (0.002906814683228731, 610.4310913085938)
                                (0.0029427013359963894, 617.96728515625)
                                (0.002978587755933404, 625.50341796875)
                                (0.003014474408701062, 633.0396118164062)
                                (0.0030503610614687204, 640.5758056640625)
                                (0.0030862477142363787, 648.1119995117188)
                                (0.0031221341341733932, 655.648193359375)
                                (0.0031580207869410515, 663.1843872070312)
                                (0.0031939074397087097, 670.7205810546875)
                                (0.003229794092476368, 678.2567749023438)
                                (0.0032656805124133825, 685.7929077148438)
                                (0.0033015671651810408, 693.3291015625)
                                (0.003337453817948699, 700.8652954101562)
                                (0.0033733404707163572, 708.4014892578125)
                                (0.003409226890653372, 715.9376220703125)
                                (0.0034451137762516737, 723.473876953125)
                                (0.0034810001961886883, 731.0100708007812)
                                (0.0035168868489563465, 738.5462646484375)
                                (0.0035527735017240047, 746.0824584960938)
            };
                \addlegendentry{Reference}

                \addplot[color=black, dashed, mark=o, mark
                options=solid] 
                coordinates {
                    (0.0, 5.7147216796875)
                    (3.5886652767658234e-05, -23.735107421875)
                        (7.177330553531647e-05, 11.71942138671875)
                        (0.00010765972547233105, 23.421173095703125)
                    (0.00014354661107063293, 37.663787841796875)
                    (0.00017943303100764751, 30.441619873046875)
                    (0.0002153199166059494, 39.578460693359375)
                    (0.000251206336542964, 64.08023071289062)
                    (0.00028709275647997856, 58.8199462890625)
                    (0.0003229794092476368, 72.88702392578125)
                    (0.00035886606201529503, 84.26846313476562)
                    (0.00039475271478295326, 79.57620239257812)
                    (0.00043063913471996784, 100.57846069335938)
                    (0.0004665257874876261, 108.44412231445312)
                    (0.0005024124402552843, 107.71923828125)
                    (0.0005382990930229425, 110.16159057617188)
                    (0.0005741857457906008, 108.14398193359375)
                    (0.0006100721657276154, 110.28070068359375)
                    (0.0006459588184952736, 141.64248657226562)
                    (0.0006818454712629318, 148.55902099609375)
                    (0.0007177320076152682, 153.25430297851562)
                    (0.0007536186603829265, 168.27151489257812)
                    (0.0007895051967352629, 170.11703491210938)
                    (0.0008253918495029211, 170.20172119140625)
                    (0.0008612785022705793, 169.92596435546875)
                    (0.0008971650386229157, 187.90673828125)
                    (0.0009330515749752522, 209.11984252929688)
                    (0.0009689382277429104, 202.5264892578125)
                    (0.0010048248805105686, 209.69589233398438)
                    (0.001040711416862905, 230.046630859375)
                    (0.0010765980696305633, 233.22817993164062)
                    (0.0011124846059828997, 233.3114013671875)
                    (0.001148371258750558, 245.0931396484375)
                    (0.0011842579115182161, 261.03619384765625)
                    (0.0012201444478705525, 256.9845275878906)
                    (0.0012560311006382108, 263.00531005859375)
                    (0.001291917753405869, 250.47042846679688)
                    (0.0013278044061735272, 270.0738525390625)
                    (0.0013636908261105418, 283.39727783203125)
                    (0.0013995774788782, 285.8508605957031)
                    (0.0014354640152305365, 310.14337158203125)
                    (0.0014713506679981947, 306.907958984375)
                    (0.001507237204350531, 310.07666015625)
                    (0.0015431238571181893, 306.9996337890625)
                    (0.0015790103934705257, 313.355712890625)
                    (0.001614897046238184, 327.7598876953125)
                    (0.0016507835825905204, 357.76239013671875)
                    (0.0016866702353581786, 360.17645263671875)
                    (0.0017225568881258368, 359.3791809082031)
                    (0.001758443540893495, 372.453125)
                    (0.0017943300772458315, 375.631591796875)
                    (0.0018302167300134897, 376.362548828125)
                    (0.0018661032663658261, 394.2818908691406)
                    (0.0019019899191334844, 398.1605224609375)
                    (0.0019378764554858208, 400.68841552734375)
                    (0.001973763108253479, 412.6317138671875)
                    (0.0020096495281904936, 422.47601318359375)
                    (0.002045536180958152, 420.0184326171875)
                    (0.00208142283372581, 440.59344482421875)
                    (0.0021173092536628246, 448.3256530761719)
                    (0.002153195906430483, 450.80322265625)
                    (0.002189082559198141, 462.7199401855469)
                    (0.0022249692119657993, 473.32574462890625)
                    (0.0022608558647334576, 482.88275146484375)
                    (0.002296742517501116, 491.8285217285156)
                    (0.002332629170268774, 498.6643371582031)
                    (0.0023685158230364323, 495.85906982421875)
                    (0.002404402242973447, 510.0483093261719)
                    (0.002440288895741105, 516.8666381835938)
                    (0.0024761755485087633, 517.1259765625)
                    (0.0025120622012764215, 514.7976684570312)
                    (0.002547948621213436, 537.4881591796875)
                    (0.0025838352739810944, 543.0405883789062)
                    (0.0026197219267487526, 551.8495483398438)
                    (0.002655608579516411, 552.8910522460938)
                    (0.0026914949994534254, 572.0118408203125)
                    (0.0027273816522210836, 576.292236328125)
                    (0.002763268304988742, 586.0355834960938)
                    (0.0027991549577564, 590.7261962890625)
                    (0.0028350413776934147, 606.9089965820312)
                    (0.002870928030461073, 603.9705810546875)
                    (0.002906814683228731, 610.593505859375)
                    (0.0029427013359963894, 604.6953125)
                    (0.002978587755933404, 625.6873168945312)
                    (0.003014474408701062, 638.7633056640625)
                    (0.0030503610614687204, 647.737548828125)
                    (0.0030862477142363787, 653.0260009765625)
                    (0.0031221341341733932, 651.3060302734375)
                    (0.0031580207869410515, 662.31884765625)
                    (0.0031939074397087097, 672.2776489257812)
                    (0.003229794092476368, 681.0496215820312)
                    (0.0032656805124133825, 692.4031982421875)
                    (0.0033015671651810408, 684.5560302734375)
                    (0.003337453817948699, 692.5762939453125)
                    (0.0033733404707163572, 717.0921630859375)
                    (0.003409226890653372, 728.9896850585938)
                    (0.0034451137762516737, 725.2518920898438)
                    (0.0034810001961886883, 730.0433349609375)
                    (0.0035168868489563465, 746.5062255859375)
                    (0.0035527735017240047, 752.0394287109375)
                };
                \addlegendentry{$d_t = 100$}
            \end{axis}
        \end{tikzpicture}
        \caption{\textbf{Elasticity - prediction in 100 time steps:} Prediction
        of the LIF from \autoref{eq:LIF} for $d_t = 100$. Fluctuations around
    the true solution can be observed.}
        \label{fig:pred_elasticity_100}
\end{figure}
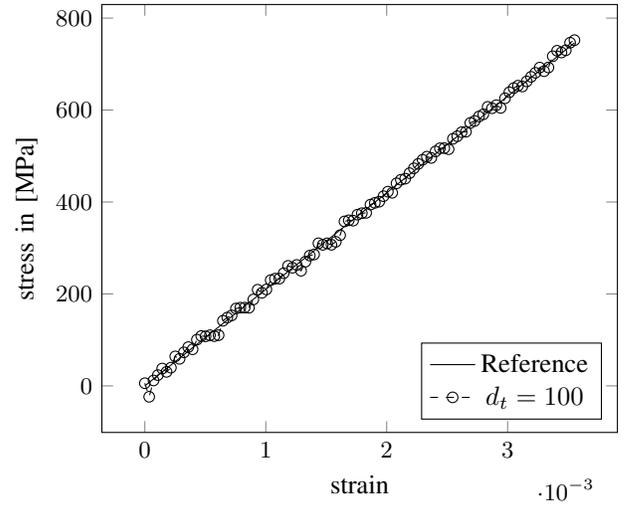

The results are illustrated in \autoref{fig:elasticity_timesteps}. It can be
seen, that the mean relative error is lowest for $d_t = 5$ time steps. For
$d_t = 2$, the error is larger. This could be caused by a lack of a sufficient
number of time steps for the neuron dynamics to effectively be calculated. It
can be understood as a
failure due to too large time steps in the explicit stepping scheme in
\autoref{eq:membrane_potential}. Clearly, the highest error can
be observed for $d_t = 100$ time steps. In contrast, as depicted in
\autoref{fig:elasticity_timesteps_end}, the error at the last time step is
lowest for $d_t = 100$ time steps. To illustrate the cause, the prediction of
the network for two different samples, one for $d_t = 5$ and one for $d_t = 100$
time steps are shown in \autoref{fig:pred_elasticity_5} and
\autoref{fig:pred_elasticity_100}, respectively. While good agreement on the
endpoints is apparent, fluctuation during the rest of the time steps causes the
rise in the error. Seemingly, the LIF has difficulties regressing a large
number of time steps. This could be caused by the lack of recurrent connections
in the LIF formulation from \autoref{eq:LIF}, where history dependency is only
weakly included in the form of the membrane potential. To counter this problem,
recurrent LIFs will be introduced in \autoref{sec:example2}.\newline

\textbf{Remark:}
The seemingly simple linear regression task provides a challenge for SNN, as
effectively an ordinary differential equation has to be fitted to a linear
function while relying on binary information transmission and inexact
gradients.

\begin{figure}[htb]
    \centering
    \begin{tikzpicture}
        \begin{axis}[ 
                xlabel={time steps $d_t$},
                ylabel={mean rel. error last time step},
            ] 

            \addplot[color=black, mark=x, mark options=solid] 
            coordinates {
                    (2, 0.20624607801437378)
                    (5, 0.16854660212993622)
                    (10, 0.21536597609519958)
                    (20, 0.06671145558357239)
                    (50, 0.05876411125063896)
                    (100, 0.05478934943675995)
            };
        \end{axis}
    \end{tikzpicture}
    \caption{\textbf{Elasticity - error at last time step:}  Mean relative error
    for the last time step with respect to the different total number of time
steps. The error is converging for a larger number of time steps.}
    \label{fig:elasticity_timesteps_end} 
\end{figure}
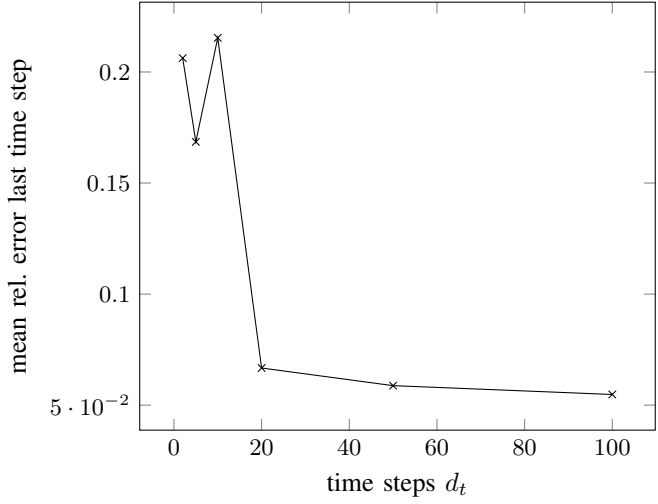

\begin{figure}[htb]
    \centering
    \begin{tikzpicture}
        \begin{axis}[ 
                xlabel={strain},
                ylabel={stress in [MPa]},
                legend pos=south east,
            ] 

            \addplot[color=black] 
            coordinates {
                                    (2.3283064365386963e-10, 0.0)
                                    (0.0007301522418856621, 153.33197021484375)
                                    (0.001460304600186646, 306.6639709472656)
                                    (0.00219045695848763, 459.9959716796875)
                                    (0.002920609200373292, 613.3279418945312)
            };
                \addlegendentry{Reference}
                \addplot[color=black, dotted, mark=o, mark options=solid] 
            coordinates {
                                    (2.3283064365386963e-10, 0.75775146484375)
                                    (0.0007301522418856621, 154.013916015625)
                                    (0.001460304600186646, 320.5915222167969)
                                    (0.00219045695848763, 464.82666015625)
                                    (0.002920609200373292, 608.4132690429688)

            };
                \addlegendentry{$d_t = 5$}
            \end{axis}
        \end{tikzpicture}
        \caption{\textbf{Elasticity - prediction in 5 time steps:} Prediction of
        the LIF from \autoref{eq:LIF} for $d_t = 5$. Deviations from the true
    solution can be observed in the middle part.}
        \label{fig:pred_elasticity_5}
\end{figure}
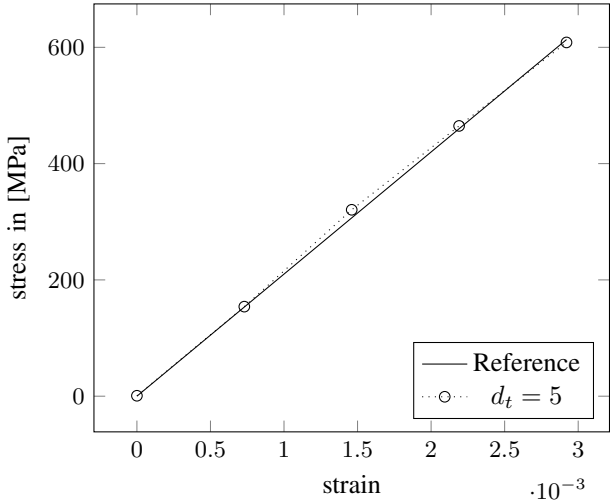

\section{Nonlinear regression using RLIF} 
\noindent In order to counter the problems of vanishing information for a large
number of time steps encountered in the preceding section, a recurrent SNN
architecture is proposed (\autoref{sec:rlif}). Its performance is demonstrated
by means of a numerical example in \autoref{sec:ramberg_osgood}.

\label{sec:example2}\noindent
\subsection{Recurrent Leaky Integrate and Fire (RLIF)} 
\label{sec:rlif} \noindent
The standard LIF is a feed-forward neuron, such that information is flowing
unidirectionally in the form of spikes. By adding a feedback loop, a
\textit{recurrent LIF} (RLIF) can be formulated, which builds on the standard
recurrent neural network (RNN) formulation. This enables the network to use
relationships along several time steps for the prediction of the current time
step. It was shown in \cite{pascanu2013difficulty}, that recurrent loops can
retain information for a relatively longer number of time steps when compared to
their non-recurrent counterparts.

Here, the formulation of the hidden layer in \autoref{eq:hidden_t} includes
additional recurrent weights $\bm{V}^{(l)}_{\eta}$, such that
\begin{equation} 
    h_{\eta, t}^{(l)} = \phi^{(l)}\left(\bm{W}^{(l)}_{\eta} \bm{h}_t^{(l-1)} +
    \bm{V}^{(l)}_{\eta} \bm{h}_{t - 1}^{(l-1)}\right).
    \label{eq:hidden_rnn} 
\end{equation} 
In this RNN, the influence of the preceding time step is explicitly included by
means of additional \textit{recurrent weights} $\bm{V}^{(l)}_{\eta}$. The
 resulting set of trainable parameters reads
\begin{equation}
    \bm{\theta}_{\text{RNN}}=\left\{\bm{W}_{\eta}^{(l)},
    \bm{V}^{(l)}_{\eta}\right\}.
    \label{eq:parameters_rnn}
\end{equation} 
The RNN formulation can be included in the LIF formulation from
\autoref{eq:membrane_potential} to obtain an RLIF, such that
\begin{align}
    \label{eq:membrane_potential_rlif}
    U_{\eta, t}^{(l)} &= \beta_{\eta}^{(l)} U_{\eta, t - 1}^{(l)} +
    \bm{W}^{(l)}_{\eta} \bm{h}_{t}^{(l-1)} \\
    &\quad+ \bm{V}^{(l)}_{\eta} \bm{h}_{t - 1}^{(l-1)}
    - \phi_{\text{spk}, t - 1}^{(l)}
    U_{\text{thr}, \eta}^{(l)}, \nonumber
\end{align}
where $U_{\eta, t}^{(l)}$ is again the membrane potential of the $\eta$-th
neural unit at time $t$, $U_{\text{thr}, \eta}^{(l)}$ denotes the membrane
threshold, $\beta_\eta^{(l)}$ is the membrane potential decay rate and
$\bm{W}^{(l)}_{\eta} \bm{h}_{t}^{(l-1)}$ is the standard ANN weight multiplied
with the preceding layer at the current time step, respectively. Additionally,
$\bm{V}^{(l)}_{\eta}$ denotes the recurrent weights from
\autoref{eq:hidden_rnn}. This leads to the following set of trainable parameters
\begin{equation}
    \bm{\theta}_{\text{RLIF}}=\left\{\bm{W}_{\eta}^{(l)},
    \bm{V}^{(l)}_{\eta}, \beta_{\eta}^{(l)},
    U_{\text{thr}, \eta}^{(l)}\right\}.
    \label{eq:parameters_rlif}
\end{equation} 

\subsection{Numerical experiment: Ramberg-Osgood}
\label{sec:ramberg_osgood}
\begin{figure}[htb]
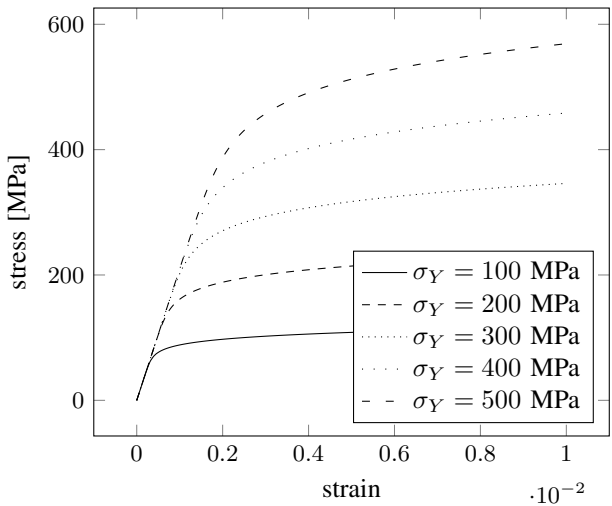

    \centering

    \caption{\textbf{Ramberg-Osgood - reference solutions.} Stress-strain curves
    of the Ramberg-Osgood material model for five different values of the yield
stress $\sigma_Y$ obtained with Newton-Raphson algorithm.}
    \label{fig:ramberg_osgood} 
\end{figure}

\noindent The performance of the RLIF is investigated toward nonlinear function
regression. The well-known nonlinear Ramberg-Osgood power law for modeling 
history-independent plasticity is chosen. The formulation of the stress $\sigma$
with respect to the strain $\varepsilon$ reads
\begin{equation}
    \varepsilon = \frac{\sigma}{E} + 0.002 \left(\frac{\sigma}{\sigma_Y}
    \right)^{n},
    \label{eq:ramberg_osgood}
\end{equation}
where $\varepsilon$ is the infinitesimal, one-dimensional elastic strain,
$\sigma$ denotes the one-dimensional Cauchy stress, $E$ is Young's modulus,
$\alpha$ and $n$ are constants describing the hardening behavior of plastic
deformation and $\sigma_Y$ is the yield strength of the material. In
\autoref{fig:ramberg_osgood}, different stress-strain curves are depicted for
different yield strength values, obtained with a classical Newton-Raphson
method. Note that this plasticity model is only suited for a single
loading direction and does not incorporate accumulation of plastic strain. It is
only used as a prototypical nonlinear model to show the ability of the RLIF to
regress over a moderate number of time steps.

To this end, the general model described in \autoref{sec:topology}
using RLIF defined in \autoref{eq:membrane_potential_rlif} is used, resulting in
the following architecture
\begin{align} 
    \label{eq:RLIF} 
    \mathcal{S} \left( \sigma_Y \right) &= 
    \bm{h}_t^{\text{pop}} \circ
    \bm{h}_t^{\text{dec}} \circ
    \bm{h}_t^{\text{RLIF}} \circ 
    \bm{h}_t^{\text{RLIF}} \circ 
    \bm{h}_t^{\text{RLIF}} \\ &\quad\circ 
    \bm{h}_t^{\text{const}} \left( \sigma_Y \right) 
    = \sigma_t, \nonumber
\end{align} 
where the yield strength $\sigma_Y$ is provided as a constant current, that is a
constant spike train, for each time step $d_t$. 

The training data consists of yield strength $\sigma_Y$ as input for fixed
strains in the interval $\varepsilon = [0, 0.01]$ for $d_t = 20$ time steps. The
yield strength is uniformly sampled in the interval $\sigma_Y = [100, 500]$ MPa,
and the stress output is calculated according to \autoref{eq:ramberg_osgood}.
The Young's modulus is chosen as $E = 2.1 \times 10^5$ MPa and $n = 10$. Three
datasets are generated, namely a training set, a validation and test set with
$n_{\text{train}} = n_{\text{val}} = n_{\text{test}} = 1024$ samples,
respectively. All three sets are standardized using the mean and standard
deviation from the training set. The batch size is chosen as $n_{\text{batch}} =
1024$. The number of neurons $n_u$ is chosen as $n_u = 128$ and is kept constant
over all layers. The training is carried out for $5 \times 10^3$ epochs. The
model performing best on the validation set is chosen for subsequent
evaluations. The mean relative error and the mean relative error of the last
time step with respect to the test set are reported.

\begin{figure}[htb]
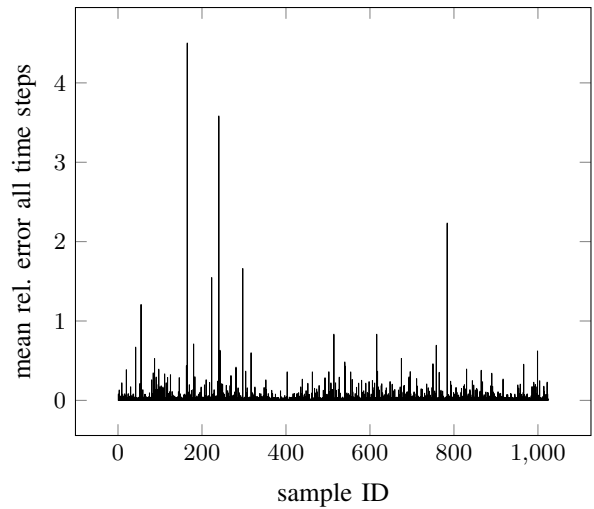

    \centering

    \caption{\textbf{Ramberg-Osgood - RLIF test error.} The mean relative error
    over all time steps for 1024 samples from the test set for the numerical
experiment described in \autoref{sec:ramberg_osgood}. It can be seen, that some
outliers have large error values, resulting in a mean error over all samples of
$8.7934 \times 10^{-2}$. Most samples have a significantly lower error.}
    \label{fig:error_samples}
\end{figure}

\begin{figure}[htb]
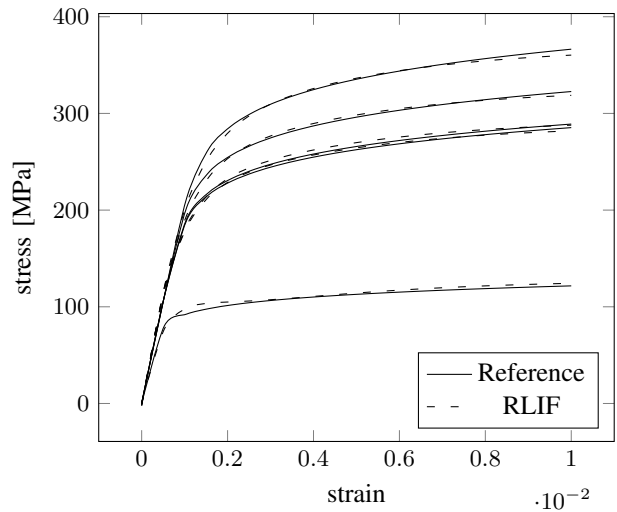

    \centering
    %
    \caption{\textbf{Ramberg-Osgood - RLIF prediction.} Prediction of the RLIF
    from \autoref{eq:RLIF} for five different yield strength $\sigma_Y$ sampled
from the test set for the nonlinear Ramberg-Osgood plasticity law. }
\label{fig:prediction_ramberg_osgood} 
\end{figure}

The results of five different samples, randomly chosen from the $n_{test}=1024$
test samples,  can be seen in \autoref{fig:prediction_ramberg_osgood}. For the
test set, a mean relative error for all time steps of {$8.7934 \times 10^{-2}$
    and a mean relative error for the last time step of $8.0200 \times 10^{-2}$
is obtained.} The predictions on these five samples are more accurate than would
be suspected from the mean relative error. The cause can be found in
\autoref{fig:error_samples}, where the mean relative error for all time steps is
plotted for every sample of the test set. It can be observed, that a small
number of samples has a much higher error than the rest, which impacts the error
measure. This is caused by the purely data-driven nature of the experiment and
can be tackled with approaches introduced in, e.g., \cite{kalina2022automated,
    masi2021thermodynamics, masi2022multiscale}. Nevertheless, the RLIF is able
    to regress on the varying yield strength $\sigma_Y$ and can predict the
    resulting nonlinear stress-strain behavior, as can be seen in the
    predictions \autoref{fig:prediction_ramberg_osgood}.  Deviations can be
    observed around the yield point as well as the endpoints of the curves. To
    be able to take into account long-term history dependent behavior, the RLIF
    formulation will be expanded towards the incorporation of explicit long-term
    memory in the next section, where a more complex plasticity model is
    investigated.

\FloatBarrier
\section{History-dependent regression using SLSTM} 
\label{sec:example3}\noindent
To extend the limited memory of the RNN in \autoref{sec:example2}, a
\textit{spiking long short-term memory neural network} (SLSTM,
\autoref{sec:slstm}) is investigated on a history-dependent plasticity model
(\autoref{sec:isotropic_hardening}).

\subsection{Spiking long short-term memory network (SLSTM)}\label{sec:slstm}
\noindent A SLSTM is the spiking version of the
standard LSTM \cite{hochreiter1997long}, where the latter is defined as
\begin{equation}
    h_{\eta, t}^{(l)} = o_{\eta, t}^{(l)} \odot \phi_{\text{tanh}} \left(
    c_{\eta, t}^{(l)}\right),
    \label{eq:lstm}
\end{equation}
with
\begin{align}
    o_{\eta, t}^{(l)} &= \phi_{\text{sigmoid}} \left(\bm{W}^{(l)}_{o, \eta}
    \bm{h}_t^{(l-1)} + \bm{V}^{(l)}_{o, \eta} \bm{h}_{t - 1}^{(l-1)}\right), \nonumber \\
    c_{\eta, t}^{(l)} &= f_{\eta, t}^{(l)} \odot c_{\eta, t-1}^{(l)} +
    i_{\eta, t}^{(l)} \odot \tilde{c}_{\eta, t}^{(l)},
    \nonumber \\
    f_{\eta, t}^{(l)} &= \phi_{\text{sigmoid}} \left(\bm{W}^{(l)}_{f, \eta}
    \bm{h}_t^{(l-1)} + \bm{V}^{(l)}_{f, \eta} \bm{h}_{t - 1}^{(l-1)}\right),
    \nonumber \\
    i_{\eta, t}^{(l)} &= \phi_{\text{sigmoid}} \left(\bm{W}^{(l)}_{i, \eta}
    \bm{h}_t^{(l-1)} + \bm{V}^{(l)}_{i, \eta} \bm{h}_{t - 1}^{(l-1)}\right),
    \nonumber \\
    \tilde{c}_{\eta, t}^{(l)} &= \phi_{\text{tanh}} \left(\bm{W}^{(l)}_{c, \eta}
    \bm{h}_t^{(l-1)} + \bm{V}^{(l)}_{c, \eta} \bm{h}_{t - 1}^{(l-1)}\right),
    \nonumber
\end{align}
where $f_t$ denotes the \textit{forget gate} with \textit{sigmoid activation}
$\phi_{\text{sigmoid}}$ or \textit{tangent hyperbolicus activation}
$\phi_{\text{tanh}}$ and corresponding weights $\bm{W}_f, \bm{V}_f$ with
absorbed biases.  The same nomenclature holds for the \textit{input gate} $i_t$,
the \textit{output gate} $o_t$, the \textit{cell input} $\tilde{c}_t$ and the
\textit{cell state} $c_t$ with their respective activations and weights.  The
new cell state $c_t$ and the output of the LSTM $h_t$ are formed using the
Hadamard or point-wise product $\odot$. The parameters of the LSTM are its
weights, such that
\begin{align} 
    \label{eq:parameters_lstm}
    \bm{\theta}_{\text{LSTM}} = \Big\{
        &\bm{W}^{(l)}_{f, \eta}, \bm{W}^{(l)}_{i, \eta}, \bm{W}^{(l)}_{o, \eta},
        \bm{W}^{(l)}_{c, \eta}, \\
        &\bm{V}^{(l)}_{f, \eta}, \bm{V}^{(l)}_{i, \eta}, 
        \bm{V}^{(l)}_{o, \eta}, \bm{V}^{(l)}_{c, \eta}.
    \Big\}. \nonumber
\end{align} 
For detailed derivations and explanations of standard LSTM, see, e.g.,
\cite{goodfellow2016deep, aggarwal2018neural}.
The SLSTM can be obtained from the LSTM by using spike activations within the
LSTM formulation from \autoref{eq:lstm}, such that
\begin{equation}
    h_{\eta, t}^{(l)} = o_{\eta, t}^{(l)} \odot \phi_{\text{tanh}} \left(
    c_{\eta, t}^{(l)}\right)  {-
    \phi_{\text{spk}, t - 1}^{(l)} U_{\text{thr}, \eta}^{(l)}},
    \label{eq:slstm}
\end{equation}
where the output $h_{\eta, t}^{(l)}$ is used to determine if a spike is produced
\begin{align}
    \phi_{\text{spk}, t}^{(l)} = 
    \begin{cases} 
        1, \quad h_{\eta, t}^{(l)} \geq U_{\text{thr}, \eta}^{(l)} \\ 
        0, \quad h_{\eta, t}^{(l)} < U_{\text{thr}, \eta}^{(l)}.
    \end{cases} 
    \label{eq:slstm_spike} 
\end{align}
In other words, the output of $h_{\eta, t}^{(l)}$ can be interpreted as the
membrane potential of the SLSTM, such that $h_{\eta, t}^{(l)} = U_{\eta,
t}^{(l)}$. A decay parameter $\beta$ is not used in this formulation. Rather
than using decay to remove information from the cell state $c_{\eta, t}^{(l)}$,
this is achieved by carefully regulated gates.

The corresponding optimization parameters of the SLSTM are 
\begin{align} 
    \label{eq:parameters_slstm} 
    \bm{\theta}_{\text{SLSTM}} = \Big\{
        &\bm{W}^{(l)}_{f, \eta}, \bm{W}^{(l)}_{i, \eta}, \bm{W}^{(l)}_{o, \eta},
        \bm{W}^{(l)}_{c, \eta}, \\
        &\bm{V}^{(l)}_{f, \eta}, \bm{V}^{(l)}_{i, \eta},
        \bm{V}^{(l)}_{o, \eta}, \bm{V}^{(l)}_{c, \eta}, U_{\text{thr},
        \eta}^{(l)} \Big\}.     \nonumber
\end{align}
Basically, the cell state $c_{\eta, t}^{(l)}$ acts as long-term memory, just
like in the standard LSTM formulation. The communication between layers is
handled via spike trains that depend on the membrane potential $h_{\eta,
t}^{(l)} = U_{\eta, t}^{(l)}$ in \autoref{eq:slstm} and the activation function
$\phi_{\text{spk}, t}^{(l)}$ from \autoref{eq:slstm_spike}.

\subsection{Numerical experiment: Isotropic hardening using SLSTM}
\label{sec:isotropic_hardening}
\begin{figure}[htb]
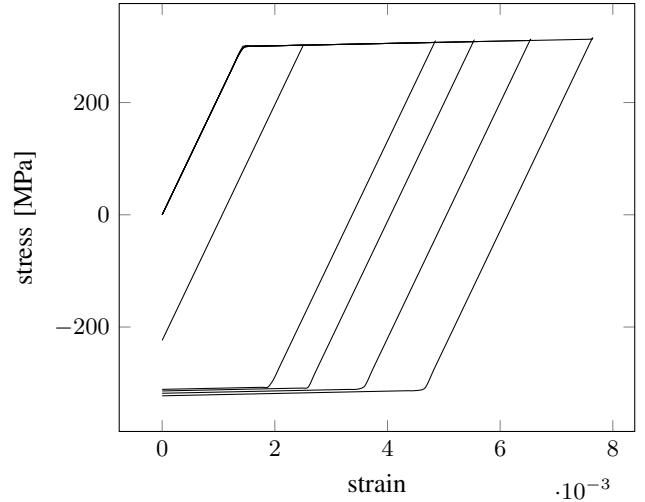

    \centering

    \caption{\textbf{Isotropic hardening - reference solutions.} Five
    stress-strain curves sampled from the isotropic hardening material model for
different maximum strains obtained from \autoref{eq:plasticity}.}
    \label{fig:plasticity} 
\end{figure}

\noindent The following numerical experiments aim to investigate the performance
of the proposed SLSTM on nonlinear, history-dependent problems. Therefore, a
one-dimensional plasticity model with isotropic hardening is investigated.
Following \cite{simo2006computational}, the model is defined by
\begin{align}
    \label{eq:plasticity}
    &1.\quad \varepsilon = \varepsilon_{\text{el}} + \varepsilon_{\text{pl}}, \\
    &2.\quad \sigma = E \left( \varepsilon - \varepsilon_{\text{pl}} \right),
    \nonumber \\
    &3.\quad \dot{\varepsilon}_{\text{pl}} = \gamma \operatorname{sign} \left(
    \sigma \right),\quad \dot{\alpha} = \gamma, \nonumber \\
    &4.\quad f \left( \sigma, \alpha \right) = | \sigma | - \left( \sigma_Y + K
    \alpha \right) \leq 0, \nonumber \\
    &5.\quad \gamma \geq 0, \quad f \left( \sigma, \alpha \right) \leq 0, \quad
    \gamma f \left( \sigma, \alpha \right) = 0, \nonumber \\
    &6.\quad \gamma \dot{f} \left( \sigma, \alpha \right) = 0, \quad \text{if}
    \; f
    \left( \sigma, \alpha \right) = 0, \nonumber
\end{align}
where
\begin{enumerate}
\item is the additive elasto-plastic split of the small-strain tensor
$\varepsilon$ into a purely elastic part $\varepsilon_{\text{el}}$ and a purely
plastic part
$\varepsilon_{\text{el}}$. 
\item denotes the elastic stress-strain
relationship for the Cauchy stress tensor $\sigma$ and elastic modulus $E$. 
\item
describes the flow rule and isotropic hardening law with consistency parameter 
$\gamma$ and equivalent plastic strain $\alpha$. 
\item gives the yield condition
$f \left( \sigma, \alpha \right)$ with hardening modulus $K$. \item denotes the
Kuhn-Tucker complementarity conditions and 
\item describes the consistency
condition. 
\end{enumerate}
In \autoref{fig:plasticity}, different stress-strain paths are shown for varying
strains. Especially long-time dependencies are of interest. To this end, the
predictive capabilities of the SNN are investigated for inference over $d_t =
100$ time steps, where the elasto-plastic model is evaluated using a classical
explicit return-mapping algorithm, see \cite{simo2006computational}.

The training data consists of strain as input, uniformly sampled in the interval
$\varepsilon = [0, 0.01]$, and stress as output calculated according to
\autoref{eq:plasticity}. The yield stress is chosen as $\sigma_Y = 300$ MPa, the
elastic modulus $E = 2.1 \times 10^5$ MPa and the hardening modulus as $2.1
\times 10^4$ MPa. Three datasets are generated, namely a training set with
$n_{\text{train}} = 10240$ samples and a validation and test set with
$n_{\text{val}} = n_{\text{test}} = 1024$ samples, respectively. All three sets
are standardized using the mean and standard deviation from the training set.
The batch size is chosen as $n_{\text{batch}} = 1024$. The training is carried
out for 500 epochs. The model performing best on the validation set is chosen
for subsequent evaluations. The mean relative error accumulated over all time
steps and the mean relative error of the last time step with respect to the test
set are reported. The last time step is of special importance, as in the case of
numerical simulations, only the resulting stress from the last time step is used
for subsequent calculations.

\begin{figure}[htb]
    \centering
    \begin{tikzpicture}
        \begin{axis}[ 
                xlabel={number of hidden neurons $n_u$},
                ylabel={mean rel. error all time steps},
                legend pos=north east,
                xticklabels={ 
                    {$16$}, 
                    {$32$}, 
                    {$64$}, 
                    {$128$},
                    {$256$},
                },
                xtick={1,...,5},
            ] 

            \addplot[color=black, mark=x, mark options=solid] 
            coordinates {
                (1, 0.16785261034965515)
                (2, 0.12519752979278564)
                (3, 0.06794638931751251)
                (4, 0.07605953514575958)
                (5, 0.054365772753953934)

            };
            \addlegendentry{Output: 16}

            \addplot[color=black, dotted, mark=triangle, mark options=solid] 
            coordinates {
                (1, 0.1700017899274826)
                (2, 0.07994617521762848)
                (3, 0.06786447763442993)
                (4, 0.05668754130601883)
                (5, 0.055632948875427246)
            };
            \addlegendentry{Output: 32}

            \addplot[color=black, dashed, mark=o, mark options=solid] 
            coordinates {
                (1, 0.1550946682691574)
                (2, 0.11850263923406601)
                (3, 0.06786084175109863)
                (4, 0.0674203410744667)
                (5, 0.05244522914290428)
            };
            \addlegendentry{Output: 64}

            \addplot[color=black, dashdotted, mark=diamond, mark options=solid] 
            coordinates {
                (1, 0.1916196644306183)
                (2, 0.07717157155275345)
                (3, 0.0775095671415329)
                (4, 0.07750254124403)
                (5, 0.0666898787021637)
            };
            \addlegendentry{Output: 128}

            \addplot[color=black, loosely dotted, mark=square, mark
            options=solid] 
            coordinates {
                (1, 0.2214195877313614)
                (2, 0.104187972843647)
                (3, 0.05959603190422058)
                (4, 0.07958905398845673)
                (5, 0.07538670301437378)
            };
            \addlegendentry{Output: 256}

        \end{axis}
    \end{tikzpicture}
    \caption{\textbf{Isotropic hardening - error versus width.} The mean
        relative error of the last time step versus the number of hidden neurons
        per layer is shown for different numbers of output neurons in the
        isotropic hardening experiment from \autoref{sec:example3} using the
        SLSTM from \autoref{eq:slstm_topology}.}
    \label{fig:convergence} 
\end{figure}
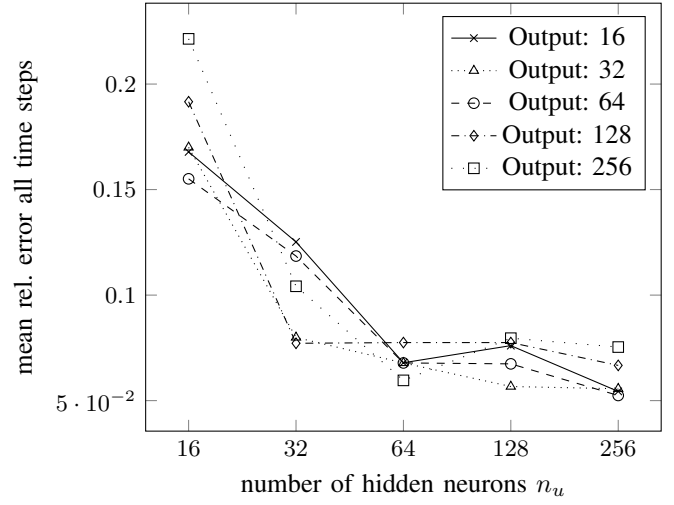

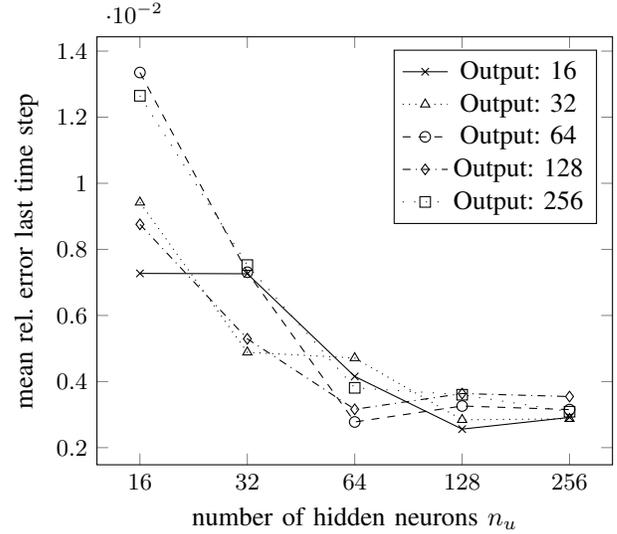
\begin{figure}[htb]
    \centering
    \begin{tikzpicture}
        \begin{axis}[ 
                xlabel={number of hidden neurons $n_u$},
                ylabel={mean rel. error last time step},
                legend pos=north east,
                xticklabels={ 
                    {$16$}, 
                    {$32$}, 
                    {$64$}, 
                    {$128$},
                    {$256$},
                },
                xtick={1,...,5},
            ] 

            \addplot[color=black, mark=x, mark options=solid] 
            coordinates {
                (1, 0.007272565271705389)
                (2, 0.007262129802256823)
                (3, 0.004154578782618046)
                (4, 0.0025608548894524574)
                (5, 0.0029142536222934723)
            };
            \addlegendentry{Output: 16}

            \addplot[color=black, dotted, mark=triangle, mark options=solid] 
            coordinates {
                (1, 0.009422734379768372)
                (2, 0.004885466769337654)
                (3, 0.004713095258921385)
                (4, 0.002842646325007081)
                (5, 0.002872937824577093)
            };
            \addlegendentry{Output: 32}

            \addplot[color=black, dashed, mark=o, mark options=solid] 
            coordinates {
                (1, 0.013352872803807259)
                (2, 0.007308153435587883)
                (3, 0.0027775838971138)
                (4, 0.0032596453092992306)
                (5, 0.003152076620608568)
            };
            \addlegendentry{Output: 64}

            \addplot[color=black, dashdotted, mark=diamond, mark options=solid] 
            coordinates {
                (1, 0.00875802431255579)
                (2, 0.00529337115585804)
                (3, 0.0031600911170244217)
                (4, 0.0036365645937621593)
                (5, 0.003546477062627673)
            };
            \addlegendentry{Output: 128}

            \addplot[color=black, loosely dotted, mark=square, mark
            options=solid] 
            coordinates {
                (1, 0.012647812254726887)
                (2, 0.007524486165493727)
                (3, 0.0038105123676359653)
                (4, 0.0036011699121445417)
                (5, 0.0030870940536260605)
            };
            \addlegendentry{Output: 256}

        \end{axis}
    \end{tikzpicture}
    \caption{\textbf{Isotropic hardening - error versus width.} The mean
        relative error of all time steps versus the number of hidden neurons per
        layer is shown for different numbers of output neurons in the isotropic
        hardening experiment from \autoref{sec:example3} using the SLSTM from
    \autoref{eq:slstm_topology}.}
    \label{fig:convergence_end} 
\end{figure}

The first study investigates the prediction accuracy as a function of (1) the
number of output neurons, which participate in the population regression
outlined in \autoref{sec:topology} and (2) different capacities of the SLSTM in
the sense of layer width. To this end, the SLSTM defined in
\autoref{eq:slstm_spike} is used, resulting in the following architecture:
\begin{align} 
    \label{eq:slstm_topology} 
    \mathcal{S}_{\textit{SLSTM}} \left( \varepsilon_t \right) &= 
    \bm{h}_t^{\text{pop}} \circ
    \bm{h}_t^{\text{dec}} \circ
    \bm{h}_t^{\text{SLSTM}} \circ 
    \bm{h}_t^{\text{SLSTM}} \circ 
    \bm{h}_t^{\text{SLSTM}} \\ &\quad\circ 
    \bm{h}_t^{\text{const}} \left( \varepsilon_t \right)
    = \sigma_t.  \nonumber
\end{align} 
Multiple simulations with output neurons and hidden layers drawn from the grid
$n_u \times n_o = [16, 32, 64, 128, 256] \times [16, 32, 64, 128, 256]$ are
carried out. The resulting mean relative error for all time steps with respect
to the test set is shown in \autoref{fig:convergence}, whereas the resulting
mean relative error of the last time step with respect to the test set is
depicted in \autoref{fig:convergence_end}. A clear convergence behavior can be
observed for the number of hidden neurons $n_u$, where larger numbers of neurons
lead to lower errors. For the number of output neurons $n_o$, a tendency can be
observed upon convergence with respect to $n_u$. For the largest number of
hidden neurons $n_u = 256$, the mean relative error over all time steps and the
mean relative error of the last time step get larger for $n_o = [128, 256]$
output neurons, whereas for $n_o = [16, 32, 64]$ the errors are almost the same.
The lowest mean relative error for all time steps is {$5.2445 \times 10^{-2}$}
for $n_u = 256$ hidden neurons per layer and $n_o = 64$ output neurons. The
lowest mean relative error for the last time steps is $2.8729 \times 10^{-3}$
$n_u = 256$ hidden neurons per layer and $n_o = 32$ output neurons. Again, the
seemingly high errors are caused by outliers polluting the average, as described
in \autoref{sec:ramberg_osgood}. The same counter-measures can be applied to
prohibit outliers, e.g., by enforcing thermodynamic consistency.

\begin{figure}[htb]
    \centering
    \begin{tikzpicture}
        \begin{axis}[ 
                xlabel={strain},
                ylabel={stress [MPa]},
                legend pos=south east,
            ] 

            \addplot[color=black] 
            coordinates {
                (0.0, 0.0)
                (9.978911839425564e-05, 20.95569610595703)
                (0.00019957800395786762, 41.91139221191406)
                (0.00029936712235212326, 62.867088317871094)
                (0.0003991562407463789, 83.82278442382812)
                (0.0004989451263099909, 104.77847290039062)
                (0.0005987343611195683, 125.73417663574219)
                (0.0006985232466831803, 146.68988037109375)
                (0.0007983122486621141, 167.64556884765625)
                (0.000898101250641048, 188.60125732421875)
                (0.0009978902526199818, 209.55694580078125)
                (0.0010976793710142374, 230.51266479492188)
                (0.0011974683729931712, 251.46835327148438)
                (0.001297257374972105, 272.4240417480469)
                (0.0013970464933663607, 293.3797607421875)
                (0.0014968354953452945, 300.1419372558594)
                (0.0015966244973242283, 300.34942626953125)
                (0.001696413499303162, 300.5569152832031)
                (0.001796202501282096, 300.7643737792969)
                (0.0018959916196763515, 300.97186279296875)
                (0.0019957805052399635, 301.1793518066406)
                (0.002095569623634219, 301.3868103027344)
                (0.002195358742028475, 301.5943298339844)
                (0.002295147627592087, 301.8017883300781)
                (0.0023949367459863424, 302.00927734375)
                (0.002494725864380598, 302.2167663574219)
                (0.00259451474994421, 302.4241943359375)
                (0.0026943038683384657, 302.6317138671875)
                (0.0027940929867327213, 302.83917236328125)
                (0.0028938818722963333, 303.04669189453125)
                (0.002993670990690589, 303.2541809082031)
                (0.0030934601090848446, 303.461669921875)
                (0.003193248761817813, 303.66912841796875)
                (0.0032930378802120686, 303.8765869140625)
                (0.0033928267657756805, 304.0841064453125)
                (0.003492615884169936, 304.29156494140625)
                (0.003592405002564192, 304.4990539550781)
                (0.003692193888127804, 304.70654296875)
                (0.0037919830065220594, 304.9140319824219)
                (0.003891772124916315, 305.1214904785156)
                (0.003991561010479927, 305.3289794921875)
                (0.004091349896043539, 305.5364685058594)
                (0.004191139247268438, 305.74395751953125)
                (0.00429092813283205, 305.9514465332031)
                (0.004390717018395662, 306.158935546875)
                (0.004490505903959274, 306.3663635253906)
                (0.004590295255184174, 306.5738830566406)
                (0.004690084140747786, 306.7813415527344)
                (0.004789873491972685, 306.9888610839844)
                (0.004889662377536297, 307.1963195800781)
                (0.004889662377536297, 307.1962890625)
                (0.004789873491972685, 286.2406311035156)
                (0.004690084140747786, 265.2848815917969)
                (0.004590295255184174, 244.3292236328125)
                (0.004490505903959274, 223.37344360351562)
                (0.004390717018395662, 202.41778564453125)
                (0.00429092813283205, 181.4621124267578)
                (0.004191139247268438, 160.50645446777344)
                (0.004091349896043539, 139.55068969726562)
                (0.003991561010479927, 118.59501647949219)
                (0.003891772124916315, 97.63935089111328)
                (0.0037919830065220594, 76.68363952636719)
                (0.003692193888127804, 55.72792053222656)
                (0.003592405002564192, 34.772254943847656)
                (0.003492615884169936, 13.816539764404297)
                (0.0033928267657756805, -7.1391754150390625)
                (0.0032930378802120686, -28.09484100341797)
                (0.003193248761817813, -49.050567626953125)
                (0.0030934601090848446, -70.00617218017578)
                (0.002993670990690589, -90.9618911743164)
                (0.0028938818722963333, -111.91761016845703)
                (0.0027940929867327213, -132.87326049804688)
                (0.0026943038683384657, -153.8289794921875)
                (0.00259451474994421, -174.78469848632812)
                (0.002494725864380598, -195.7403564453125)
                (0.0023949367459863424, -216.69607543945312)
                (0.002295147627592087, -237.65179443359375)
                (0.002195358742028475, -258.6074523925781)
                (0.002095569623634219, -279.56317138671875)
                (0.0019957805052399635, -300.5188903808594)
                (0.0018959916196763515, -307.3376770019531)
                (0.001796202501282096, -307.5451965332031)
                (0.001696413499303162, -307.7526550292969)
                (0.0015966244973242283, -307.96014404296875)
                (0.0014968354953452945, -308.1676330566406)
                (0.0013970464933663607, -308.3750915527344)
                (0.001297257374972105, -308.5826110839844)
                (0.0011974683729931712, -308.7900695800781)
                (0.0010976793710142374, -308.99755859375)
                (0.0009978902526199818, -309.2050476074219)
                (0.000898101250641048, -309.41253662109375)
                (0.0007983122486621141, -309.6199951171875)
                (0.0006985232466831803, -309.8274841308594)
                (0.0005987343611195683, -310.03497314453125)
                (0.0004989451263099909, -310.2424621582031)
                (0.0003991562407463789, -310.4499206542969)
                (0.00029936712235212326, -310.65740966796875)
                (0.00019957800395786762, -310.8648986816406)
                (9.978911839425564e-05, -311.0723876953125)
                (0.0, -311.2798767089844)
            };
            \addlegendentry{Reference}

            \addplot[color=black, dotted] 
            coordinates {
                (0.0, 0.05828857421875)
                (9.978911839425564e-05, 21.488277435302734)
                (0.00019957800395786762, 42.32157897949219)
                (0.00029936712235212326, 62.74219512939453)
                (0.0003991562407463789, 83.5887222290039)
                (0.0004989451263099909, 104.5667724609375)
                (0.0005987343611195683, 125.671142578125)
                (0.0006985232466831803, 146.39703369140625)
                (0.0007983122486621141, 167.4595947265625)
                (0.000898101250641048, 188.3531494140625)
                (0.0009978902526199818, 209.24276733398438)
                (0.0010976793710142374, 230.22683715820312)
                (0.0011974683729931712, 251.1260986328125)
                (0.001297257374972105, 272.11395263671875)
                (0.0013970464933663607, 293.02484130859375)
                (0.0014968354953452945, 300.1803894042969)
                (0.0015966244973242283, 300.4608459472656)
                (0.001696413499303162, 300.5558166503906)
                (0.001796202501282096, 300.8134765625)
                (0.0018959916196763515, 300.9676208496094)
                (0.0019957805052399635, 301.1869812011719)
                (0.002095569623634219, 301.4700927734375)
                (0.002195358742028475, 301.845947265625)
                (0.002295147627592087, 302.12200927734375)
                (0.0023949367459863424, 302.4134826660156)
                (0.002494725864380598, 302.57745361328125)
                (0.00259451474994421, 302.7739562988281)
                (0.0026943038683384657, 302.9772033691406)
                (0.0027940929867327213, 303.24951171875)
                (0.0028938818722963333, 303.5025634765625)
                (0.002993670990690589, 303.7061767578125)
                (0.0030934601090848446, 303.96649169921875)
                (0.003193248761817813, 304.07318115234375)
                (0.0032930378802120686, 304.1851501464844)
                (0.0033928267657756805, 304.37701416015625)
                (0.003492615884169936, 304.7449035644531)
                (0.003592405002564192, 304.96087646484375)
                (0.003692193888127804, 305.1372985839844)
                (0.0037919830065220594, 305.377685546875)
                (0.003891772124916315, 305.56597900390625)
                (0.003991561010479927, 305.6790466308594)
                (0.004091349896043539, 305.8473815917969)
                (0.004191139247268438, 306.1568603515625)
                (0.00429092813283205, 306.3606262207031)
                (0.004390717018395662, 306.4831848144531)
                (0.004490505903959274, 306.8247985839844)
                (0.004590295255184174, 307.0841979980469)
                (0.004690084140747786, 307.2872009277344)
                (0.004789873491972685, 307.5282287597656)
                (0.004889662377536297, 307.7452087402344)
                (0.004889662377536297, 307.1761169433594)
                (0.004789873491972685, 287.6069030761719)
                (0.004690084140747786, 266.5839538574219)
                (0.004590295255184174, 245.80963134765625)
                (0.004490505903959274, 224.88525390625)
                (0.004390717018395662, 204.00616455078125)
                (0.00429092813283205, 182.9903564453125)
                (0.004191139247268438, 162.10374450683594)
                (0.004091349896043539, 141.3197784423828)
                (0.003991561010479927, 120.48515319824219)
                (0.003891772124916315, 99.26131439208984)
                (0.0037919830065220594, 78.4606704711914)
                (0.003692193888127804, 57.42417526245117)
                (0.003592405002564192, 36.67705535888672)
                (0.003492615884169936, 15.760448455810547)
                (0.0033928267657756805, -5.5377655029296875)
                (0.0032930378802120686, -26.357421875)
                (0.003193248761817813, -46.87908172607422)
                (0.0030934601090848446, -67.7619400024414)
                (0.002993670990690589, -89.52910614013672)
                (0.0028938818722963333, -110.04546356201172)
                (0.0027940929867327213, -130.267822265625)
                (0.0026943038683384657, -151.73666381835938)
                (0.00259451474994421, -172.68685913085938)
                (0.002494725864380598, -193.51858520507812)
                (0.0023949367459863424, -213.8162841796875)
                (0.002295147627592087, -234.45480346679688)
                (0.002195358742028475, -256.05084228515625)
                (0.002095569623634219, -275.99176025390625)
                (0.0019957805052399635, -298.0222473144531)
                (0.0018959916196763515, -306.60040283203125)
                (0.001796202501282096, -306.5223083496094)
                (0.001696413499303162, -306.73516845703125)
                (0.0015966244973242283, -307.0164489746094)
                (0.0014968354953452945, -307.22967529296875)
                (0.0013970464933663607, -307.4150085449219)
                (0.001297257374972105, -307.5921630859375)
                (0.0011974683729931712, -307.7712097167969)
                (0.0010976793710142374, -307.9671325683594)
                (0.0009978902526199818, -308.181884765625)
                (0.000898101250641048, -308.4151611328125)
                (0.0007983122486621141, -308.6291198730469)
                (0.0006985232466831803, -308.8007507324219)
                (0.0005987343611195683, -308.97772216796875)
                (0.0004989451263099909, -309.1639709472656)
                (0.0003991562407463789, -309.3687744140625)
                (0.00029936712235212326, -309.6034851074219)
                (0.00019957800395786762, -309.8338928222656)
                (9.978911839425564e-05, -310.0216064453125)
                (0.0, -310.14044189453125)
            };
            \addlegendentry{LSTM}

            \addplot[color=black, dashed] 
            coordinates {
                (0.0, -0.28028106689453125)
                (9.978911839425564e-05, 15.435714721679688)
                (0.00019957800395786762, 36.59218978881836)
                (0.00029936712235212326, 63.490848541259766)
                (0.0003991562407463789, 80.72649383544922)
                (0.0004989451263099909, 101.27745819091797)
                (0.0005987343611195683, 120.91270446777344)
                (0.0006985232466831803, 142.40036010742188)
                (0.0007983122486621141, 164.58908081054688)
                (0.000898101250641048, 185.56689453125)
                (0.0009978902526199818, 207.2589111328125)
                (0.0010976793710142374, 227.991943359375)
                (0.0011974683729931712, 249.99380493164062)
                (0.001297257374972105, 270.668701171875)
                (0.0013970464933663607, 292.3913879394531)
                (0.0014968354953452945, 299.49859619140625)
                (0.0015966244973242283, 299.4819641113281)
                (0.001696413499303162, 299.90704345703125)
                (0.001796202501282096, 300.6097717285156)
                (0.0018959916196763515, 301.0274353027344)
                (0.0019957805052399635, 301.266845703125)
                (0.002095569623634219, 301.645263671875)
                (0.002195358742028475, 301.857421875)
                (0.002295147627592087, 302.05767822265625)
                (0.0023949367459863424, 302.2373352050781)
                (0.002494725864380598, 302.6621398925781)
                (0.00259451474994421, 302.4937744140625)
                (0.0026943038683384657, 302.5748291015625)
                (0.0027940929867327213, 302.62091064453125)
                (0.0028938818722963333, 302.8804016113281)
                (0.002993670990690589, 303.2016296386719)
                (0.0030934601090848446, 303.4538269042969)
                (0.003193248761817813, 303.8072204589844)
                (0.0032930378802120686, 303.8349914550781)
                (0.0033928267657756805, 304.16656494140625)
                (0.003492615884169936, 304.3003234863281)
                (0.003592405002564192, 304.5251770019531)
                (0.003692193888127804, 304.69140625)
                (0.0037919830065220594, 304.9293212890625)
                (0.003891772124916315, 305.0057067871094)
                (0.003991561010479927, 305.1875)
                (0.004091349896043539, 305.47088623046875)
                (0.004191139247268438, 305.6075439453125)
                (0.00429092813283205, 305.865966796875)
                (0.004390717018395662, 306.16131591796875)
                (0.004490505903959274, 306.00616455078125)
                (0.004590295255184174, 306.37188720703125)
                (0.004690084140747786, 307.2158508300781)
                (0.004789873491972685, 307.00146484375)
                (0.004889662377536297, 307.0293273925781)
                (0.004889662377536297, 307.53497314453125)
                (0.004789873491972685, 286.4072265625)
                (0.004690084140747786, 265.9592590332031)
                (0.004590295255184174, 244.41537475585938)
                (0.004490505903959274, 224.30276489257812)
                (0.004390717018395662, 203.3580322265625)
                (0.00429092813283205, 182.4266815185547)
                (0.004191139247268438, 160.9010467529297)
                (0.004091349896043539, 139.87982177734375)
                (0.003991561010479927, 119.03627014160156)
                (0.003891772124916315, 97.87625122070312)
                (0.0037919830065220594, 77.40636444091797)
                (0.003692193888127804, 56.352298736572266)
                (0.003592405002564192, 34.76255798339844)
                (0.003492615884169936, 13.44620132446289)
                (0.0033928267657756805, -7.809516906738281)
                (0.0032930378802120686, -28.80176544189453)
                (0.003193248761817813, -49.26713562011719)
                (0.0030934601090848446, -69.80655670166016)
                (0.002993670990690589, -90.2198257446289)
                (0.0028938818722963333, -111.1015396118164)
                (0.0027940929867327213, -131.43527221679688)
                (0.0026943038683384657, -151.97406005859375)
                (0.00259451474994421, -173.47467041015625)
                (0.002494725864380598, -194.17898559570312)
                (0.0023949367459863424, -215.7254638671875)
                (0.002295147627592087, -236.5869140625)
                (0.002195358742028475, -257.15478515625)
                (0.002095569623634219, -278.28680419921875)
                (0.0019957805052399635, -300.1311340332031)
                (0.0018959916196763515, -307.0447692871094)
                (0.001796202501282096, -307.2362976074219)
                (0.001696413499303162, -307.55914306640625)
                (0.0015966244973242283, -307.8214111328125)
                (0.0014968354953452945, -307.9447021484375)
                (0.0013970464933663607, -308.30059814453125)
                (0.001297257374972105, -308.3607177734375)
                (0.0011974683729931712, -308.820556640625)
                (0.0010976793710142374, -308.95733642578125)
                (0.0009978902526199818, -309.0390319824219)
                (0.000898101250641048, -309.23046875)
                (0.0007983122486621141, -309.3912353515625)
                (0.0006985232466831803, -309.57965087890625)
                (0.0005987343611195683, -309.82275390625)
                (0.0004989451263099909, -310.07318115234375)
                (0.0003991562407463789, -310.2508239746094)
                (0.00029936712235212326, -310.5575256347656)
                (0.00019957800395786762, -310.822021484375)
                (9.978911839425564e-05, -311.1182861328125)
                (0.0, -311.31622314453125)
            };
            \addlegendentry{SLSTM}
        \end{axis}
    \end{tikzpicture}
    \caption{\textbf{Isotropic hardening - LSTM versus SLSTM.} Prediction of a
    single load path using the return-mapping algorithm as a reference, the
standard LSTM and the spiking LSTM formulation.}
    \label{fig:plasticity_prediction} 
\end{figure}
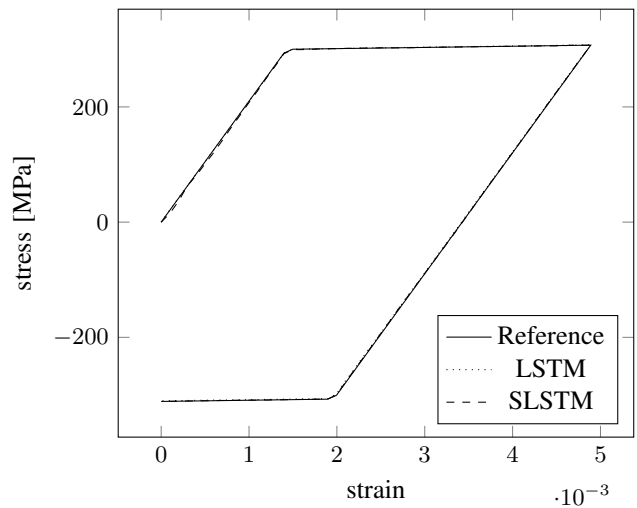

For the second experiment, the SLSTM using $n_o = 64$ output neurons and $n_u =
256$ hidden neurons per layer are compared to a standard LSTM with an equal
number of optimization parameters. The aim of this study is the comparison of
the prediction accuracy, but also the difference in memory and energy
consumption on neuromorphic hardware. For both ANN variants to be comparable,
the same topology is chosen for the LSTM as for the SLSTM, such that
\begin{align} 
    \label{eq:lstm_topology} 
    \mathcal{N}_{\textit{LSTM}} \left( \varepsilon_t \right) &= 
    \bm{h}_t^{\text{dense}} \circ
    \bm{h}_t^{\text{dense}} \circ
    \bm{h}_t^{\text{LSTM}} \circ 
    \bm{h}_t^{\text{LSTM}} \circ 
    \bm{h}_t^{\text{LSTM}} \\ &\quad\circ 
    \bm{h}_t^{\text{const}} \left( \varepsilon_t \right) = \sigma_t,
    \nonumber
\end{align} 
where the last two layers are replaced by densely connected conventional
feed-forward neural networks. Again, the training was carried out for $5 \times
10^3$ epochs and the same datasets from the previous experiments are used. The
standard LSTM from \autoref{eq:lstm_topology} reached a mean relative error of
$4.8611 \times 10^{-2}$ over all time steps and a mean relative error of $4.7569
\times 10^{-3}$ for the last time step. The SLSTM from
\autoref{eq:slstm_topology} reached a mean relative error of $9.3832 \times
10^{-2}$ over all time steps and a mean relative error of $4.0497 \times
10^{-3}$ for the last time step. The resulting prediction for one strain path is
illustrated in \autoref{fig:plasticity_prediction}. Clearly, both networks are
able to accurately predict the history-dependent, nonlinear stress-strain
behavior. 

Some deviations from the SLSTM can be seen in the beginning of the curve. The
dynamics of the spiking formulation results in a higher mean relative error over
all time steps with respect to the LSTM. However, the endpoint has a better fit
than the LSTM. This is seen in the lower error at the last time step. Whether
this is just an effect due to our experimental setting or a general feature of
the method has to be investigated in a larger statistical analysis in upcoming
studies.

To assess the potential of interfacing our model in embedded,
resource-constrained sensors in the wild, we performed a series of power
profiling experiments for our SNNs (both using LIF neurons and SLSTMs) when
processed on the Loihi neuromorphic chip \cite{davies2018loihi}. These results
are compared against their non-spiking equivalents on an NVIDIA V100 GPU. Data
were extracted using the energy profiler in \textit{KerasSpiking} v0.3.0.

The first difference in energy usage is that the spiking implementation is
measured in an `event-based’ manner, where processing only occurs when a neuron
emits a spike. In contrast, a non-spiking network processed on a GPU
continuously computes with all activations. Note that the cost of overhead did
not need to be accounted for (i.e., transferring data between devices) because
all models fit on a single device. The second difference is that SNNs require
multiple time steps of a forward pass, whereas their non-spiking counterparts do
not (unless the input to the network varies over time).

\begin{table}[htb]
    \centering
    \resizebox{0.45\textwidth}{!}{%
    \begin{tabular}{|l|c|c|l|c|c|}
    \hline \textbf{Architecture} & \multicolumn{2}{c|}{Energy}&
    \textbf{Architecture} & \multicolumn{2}{c|}{Energy} \\
    \hline Dense & Loihi (nJ) & GPU (nJ) & LSTM & Loihi (nJ) & GPU (nJ) \\
    \quad $\sbullet$ FC1 & 6.9e-2 & 0.61 & \quad $\sbullet$ LSTM1 & 0.28 & 2.5
    \\
    \quad $\sbullet$ FC2 & 1.3 & 160 & \quad $\sbullet$ LSTM2 & 11 & 2.5e3 \\
    \quad $\sbullet$ FC3 & 1.3 & 160 & \quad $\sbullet$ LSTM3 & 10 & 2.5e3 \\
    \quad $\sbullet$ FC4 & 1.2 & 160 & \quad $\sbullet$ FC1 & 2.6 & 630 \\
    \quad $\sbullet$ FC5 & 0.3 & 39 & \quad $\sbullet$ FC2 & 0.21 & 39 \\
    \hline \textbf{Total Energy} & 4.25 & 512 & \textbf{Total Energy} & 24 &
    5.7e3\\ 
    \hline\textbf{Reduction} & \multicolumn{2}{c|}{$\times$120} &
    \textbf{Reduction Factor} & \multicolumn{2}{c|}{$\times$238}\\
    \hline \textbf{Synaptic Memory} & \multicolumn{2}{c|}{0.86~MB} &
    \textbf{Synaptic Memory} & \multicolumn{2}{c|}{9.5~MB}\\
    \hline 
    \end{tabular}}
    \caption{Comparison between spiking and non-spiking forward-pass energy
    consumption and memory usage.}
    \label{tab:overhead}
\end{table}

Each network has been broken up into its constituent layers to measure how much
they contribute to energy usage on each device. The total energy consumption per
forward pass of the non-spiking network on the V100 is 512~nJ, whereas the
equivalent SNN is 4.25~nJ. This represents a 120x reduction in energy
consumption. The non-spiking LSTM network consumed 5.7~$\mu$J while the proposed
spiking-LSTM architecture required 24~nJ, a 238$\times$ reduction. Detailed
results are summarized in \autoref{tab:overhead}.

\FloatBarrier
\section{Conclusion and outlook} 
\label{sec:conclusion}\noindent
In the present study a framework for regression using SNNs
was proposed based on a membrane potential spiking decoder and a population
voting layer. Several numerical examples using different spiking neural
architectures investigated the performance of the introduced topology towards
linear, nonlinear, and history-dependent regression problems. 

First, a simple feed-forward SNN, the LIF, was derived from the classical
densely connected feed-forward ANN. It was shown, that the SNN can be seen as a
special kind of activation function, which produces binary outputs, so-called
spikes. These spikes are used to propagate information through a possibly deep
spiking neural network. The spikes occur due to the dynamic behavior of the
membrane potential inside the neuron, which rises when spikes appear at the
input and decays over time if no spikes appear. If a certain threshold value is
reached, the membrane potential is reset and the neuron emits a spike itself.
This formulation introduces more hyperparameters, which fortunately can be
learned during training. The spikes introduce sparsity in the network, which can
be effectively exploited by neuromorphic hardware to improve latency, power, and
memory efficiency. The non-differentiability of the binary spikes is
circumvented by surrogate gradients during backpropagation.

Next, a network topology was proposed, which decodes binary spikes
into real numbers, which is essential for all kinds of regression problems. A
decoding layer takes the membrane potentials of all neurons in the last spiking
layer and propagates them to a population voting layer, which provides its mean
potential resulting in a real number. The proposed topology can be used for
arbitrary temporal input and output dimensions. A simple experiment on a linear
elastic material model using LIFs showed, that the proposed topology is able to
regress the problem. It was shown, that errors are introduced for a large number
of time steps. This problem was overcome by introducing RLIF, which extends the
LIF by recurrent feedback loops. An experiment using a nonlinear Ramberg-Osgood
plasticity model showed that the proposed topology using RLIF is able to
regress varying yield limits accurately. The final extension was concerned with
the introduction of explicit long-term memories inspired by the classical LSTM
formulation, resulting in a spiking LSTM. The performance of this SLSTM was
investigated on a history-dependent isotropic hardening model, where different
load paths were accurately regressed. During prediction, the SLSTM was able to
generalize even better than the LSTM for the final load step. Furthermore,
the convergence of the proposed method was shown.

Power profiling and memory analysis were conducted on the LIF and SLSTM
networks to compare efficiency on neuromorphic hardware as against a GPU. The
Loihi neuromorphic processor was able to achieve a 120$\times$ reduction in
energy consumption when processing the dense LIF network, and the SLSTM offered
a 238$\times$ reduction in energy during inference.

{The range of possible future application scenarios enabled by regression with
Spiking Neural Networks are manifold. For instance, today's sensing systems
cannot capture all  quantities that are relevant for structural
health-monitoring. In the context of mechanics, displacement and strain are
quiet easy to assess, but the mechanical stress, which reflects the actual
response of structures and materials to deformation, remains a so-called
\textit{hidden-quantity}. Physics-informed machine learning offers the potential
to reconstruct \textit{hidden quantities} from data by leveraging information
from physical models, given in the form of partial differential equations. It is
expected that the developments in the field of neuromorphic hardware will foster
the development of a new generation of embedded systems, which will ultimately
enable  control of structures and processes based on partial differential
equations.}

\section*{Declaration of competing interest} \noindent
The authors declare that they have no known competing financial interests or
personal relationships that could have appeared to influence the work reported
in this paper.

\section*{Data availability} 
\label{sec:data}
\noindent
The code will be available upon acceptance at
\href{https://github.com/ahenkes1/HENKES_SNN}{https:github.com/ahenkes1/HENKES\_SNN} and [henkes\_code\_snn\_ZENODO].



\bibliography{literature_henkes}
\bibliographystyle{IEEEtran} 

\section{Biography Section}
 



\begin{IEEEbiographynophoto}{Alexander Henkes}
received the B.Sc. (Mechanical engineering) and M.Sc. (Mechanical engineering) degrees from University of Paderborn, Germany, in 2015 and 2018, respectively. In 2022 he received his Ph.D with honors from the Technical University of Braunschweig (TUBS), Germany. He is currently a Post-Doctoral Research Fellow at the Institute for Computational Modeling in Civil Engineering at TUBS. In 2022, he was elected as a junior member of the German Association of Applied Mathematics and Mechanics (GAMM) for his outstanding research in the field of artificial intelligence in continuum micromechanics. His research interests lies in the intersection of artificial neural networks, uncertainty quantification and continuum micromechanics.
\end{IEEEbiographynophoto}

\begin{IEEEbiographynophoto}{Jason K. Eshraghian}
(Member, IEEE) received the B.Eng. (Electrical and Electronic), L.L.B., and Ph.D. degrees from The University of Western Australia, Perth, WA, Australia, in 2017 and 2019, respectively. From 2019 to 2022, he was a Post-Doctoral Research Fellow at the University of Michigan, Ann Arbor MI, USA. He is currently an Assistant Professor with the Department of Electrical and Computer Engineering, University of California at Santa Cruz, Santa Cruz, CA, USA. His research interests include neuromorphic computing, resistive random access memory (RRAM) circuits, and spiking neural networks.
\end{IEEEbiographynophoto}

\begin{IEEEbiographynophoto}{Henning Wessels} received the B.Sc., M.Sc. and Ph.D. degrees (Mechanical engineering) from Leibniz University Hannover (LUH) in 2013, 2016 and 2019, respectively. During his Ph.D. he spent six months at the University of California, Berkeley. After a Postdoc at LUH, he has been appointed Assistant Professor (tenure track) for data-driven modeling and simulation of mechanical systems at the Technical University of Braunschweig (TUBS) in May 2021. His research aims to improve and augment physics-based numerical models from the field of computational mechanics using machine learning techniques.
\end{IEEEbiographynophoto}

\vfill

\end{document}